\documentclass{article} % For LaTeX2e
\usepackage{iclr2027_conference,times}
\usepackage{booktabs}
\usepackage{hyperref}
\usepackage{comment}
\usepackage{algpseudocode}
\usepackage{graphicx}
\usepackage{amssymb}
\usepackage{soul,color}
\usepackage{subcaption}
 
\usepackage{amsmath,amsfonts,bm}

\def\eqref#1{equation~\ref{#1}}
\def\1{\bm{1}}

\DeclareMathAlphabet{\mathsfit}{\encodingdefault}{\sfdefault}{m}{sl}
\SetMathAlphabet{\mathsfit}{bold}{\encodingdefault}{\sfdefault}{bx}{n}

\usepackage{hyperref}
\usepackage{url}

\title{Selective Backpropagation for Efficient Few-Shot Class-Incremental Learning}

\author{Eeham Khan\thanks{Corresponding author: \texttt{eeham.khan@concordia.ca}} , 
  Abdulmoumen Al-Atrash \&
  Ali Ayub \\
  PaInt Lab, Concordia University \\
  Montreal, QC, Canada \\
  \texttt{\{eeham.khan, ali.ayub\}@concordia.ca} \\
  \texttt{a.a.alatrash@outlook.com}
}

\iclrfinalcopy % Uncomment for camera-ready version, but NOT for submission.
\begin{document}

\maketitle

\begin{abstract}
Few-Shot Class-Incremental Learning (FSCIL) requires models to continuously learn new classes from limited samples while retaining prior knowledge, under strict constraints on compute and memory. Existing approaches lie along a difficult trade-off: simple fine-tuning is computationally efficient but suffers from catastrophic forgetting, replay-based methods mitigate forgetting at the cost of substantial compute and memory, and exemplar-free methods often reduce forgetting by freezing most of the backbone% or adding auxiliary optimization stages
, improving efficiency at the expense of adaptability. We propose Selective Backpropagation (SBP), a deterministic parameter budgeting framework that bridges this gap. SBP restricts gradient updates to a pre-allocated subset of network parameters, freezing past knowledge and preserving unbiased capacity for future learning, enabling rapid adaptation without %replay or 
costly mask optimization. We show that SBP achieves strong performance across standard FSCIL benchmarks while requiring training time close to that of naïve fine-tuning and substantially lower training time than prior SOTA methods. Crucially, our experiments expose a limitation of standard FSCIL evaluation: performance on short, distribution-consistent benchmarks does not necessarily predict behavior under distribution shift or over substantially longer learning horizons. We therefore evaluate FSCIL methods in cross-domain settings and over an 80-session ImageNet-1K stream. SBP remains strong across these regimes while maintaining low training cost, providing a favorable stability--plasticity--efficiency trade-off. Our code is available at \url{https://github.com/PaInt-Lab/sbp-main-public}. %Crucially, we show that methods relying on frozen backbones or rigid representations fail in cross-domain FSCIL, where distribution shift across sessions is unavoidable. In contrast, SBP remains highly adaptive and significantly outperforms all competing approaches, demonstrating an exceptionally strong stability–plasticity–efficiency balance suitable for realistic FSCIL scenarios.
\end{abstract}

\section{Introduction}

Modern deep learning systems are increasingly deployed in streaming settings where data evolves over time, new categories appear, and models must be updated repeatedly, known as continual learning (CL)~\cite{parisi2019continual, Delange_2021, mai2022online, ramasesh2020anatomy}. %A core obstacle in this regime is catastrophic forgetting (CF) \cite{goodfellow2015empiricalinvestigationcatastrophicforgetting}, where learning new data overwrites previously acquired knowledge and yields substantial performance degradation. Although research in continual learning (CL) has proposed many mitigation strategies \cite{999d99c356304cf4bf8614a6ac304332, Delange_2021, mai2022online, ramasesh2020anatomy}, the most effective approaches often rely on replay or repeated retraining, which can be prohibitively expensive when updates are frequent, models are large, or compute budgets are tight.
Within CL, common settings include Task-Incremental Learning (TIL) \cite{kirkpatrick2017overcoming, rusu2016progressive}, where task labels are known at test time, and Class-Incremental Learning (CIL) \cite{yan2021dynamically, Hou_2019_CVPR}, where task labels are unavailable at test time. Much of this literature assumes abundant labeled data when new classes arrive; an assumption that breaks in many real deployments where %new categories are rare, 
labeling is costly, and compute may be limited. We target this setting and study Few-Shot Class-Incremental Learning (FSCIL) \cite{mazumder2021few}, where each session introduces a few novel classes with only a handful of labeled examples per class.

FSCIL is a challenging problem because it combines two competing requirements: the model must remain stable on previously learned classes, i.e. mitigate forgetting without replaying any old data, while being plastic enough to incorporate novel classes from extremely limited supervision without overfitting. This tension is amplified by an \emph{accuracy--efficiency} trade-off: many methods boost accuracy by sacrificing efficiency (and vice-versa). Importantly, this dilemma cannot be assessed reliably only through accuracy on short, distribution-consistent classes. A method may preserve prior accuracy by freezing a representation that remains adequate when all classes come from the same dataset, but fail when later sessions require genuine representational adaptation. Similarly, performance over a small number of increments may not predict whether interference accumulates catastrophically over a much longer horizon. We therefore argue that standard FSCIL benchmarks are necessary for comparability but insufficient for evaluating continual learning quality: evaluation should jointly test retention, adaptation under distribution shift, behavior over longer learning horizons, and computational cost.

From a computational perspective, FSCIL methods can be broadly placed along a spectrum. At one extreme, standard fine-tuning (FT) represents the lower bound on compute: it is fast, simple, and fully adaptive, but catastrophically forgets previously learned classes. At the other extreme, replay-based or heavily regularized FSCIL methods mitigate forgetting by reintroducing stored data or auxiliary optimization stages, often incurring substantial memory and compute overhead~\cite{yoon2020xtarnet, chi2022metafscil, zhou2022few}. Between these extremes lie exemplar-free methods that reduce forgetting by freezing most of the backbone or restricting updates~\cite{zhou2022forward,peng2022few}, typically improving efficiency but sacrificing plasticity. A promising compromise is to control interference through capacity allocation: subnetwork and parameter-isolation methods restrict learning to subsets of parameters, preserving prior knowledge while still permitting adaptation \cite{mallya2018packnet,mallya2018piggyback}. However, most of these methods are developed for TIL, implicitly relying on task identity at test time, unavailable in FSCIL% and particularly problematic under severe class imbalance 
~\cite{tao2020fewshotclassincrementallearning,kang2024continual}. Furthermore, these methods incur additional costs to search for important network parameters and learn importance masks. %While a few parameter masking FSCIL methods have been developed~\cite{softnet}, %Moreover, existing FSCIL subnetwork methods 
%these methods often depend on \emph{learned} %soft 
%masks or search procedures, which adds optimization cost and hyperparameter sensitivity.
An ideal FSCIL method would combine the speed and adaptability of fine-tuning with the stability of more complex continual learning approaches, without requiring replay, learned masks, or expensive meta-training. Despite extensive prior work, this regime remains largely unexplored: existing high-performing FSCIL methods almost universally trade efficiency or adaptability for stability.

In this work, we propose %\textbf{Selective Backpropagation (SBP)}, 
a novel FSCIL framework that balances stability and efficiency under strict exemplar-free constraints. We introduce selective backpropagation (SBP) which uses a three-way budgeting scheme that partitions the network parameters into: (i) \emph{frozen} parameters that preserve previously learned knowledge, (ii) \emph{assigned} parameters for learning the current task, and (iii) \emph{free} parameters reserved for future tasks, randomly reinitialized every epoch. Crucially, SBP pre-allocates future capacity without letting current training implicitly shape it: free parameters are unallocated and receive no gradient updates, and are also reinitialized at the beginning of each epoch, so later sessions start from unbiased capacity and earlier tasks do not depend on parameters needed later. Unlike prior parameter masking methods, SBP's deterministic budget schedule eliminates mask-learning overhead and keeps per-session updates consistently cheap. 
% To avoid any representation drift in new sessions, we further introduce partial knowledge distillation over the frozen representation space. 
Additionally, we couple SBP with prototype-based weight imprinting and classification, providing a stable and geometry-consistent initialization for novel-class weights without heavy optimization. Across standard FSCIL benchmarks, SBP achieves state-of-the-art performance on CIFAR-100 and CUB-200 while remaining competitive on MiniImageNet at substantially lower training cost than several competing baselines. Additionally, experiments on cross-domain and long-horizon FSCIL benchmarks show that methods relying on frozen backbones or rigid representations %fail in this setting where a distribution shift across sessions is unavoidable
fail in settings with a stronger distribution shift and over long-horizon tasks. In contrast, SBP remains highly adaptive and outperforms competing approaches with improved training efficiency, demonstrating an exceptionally strong stability–plasticity–efficiency balance suitable for realistic FSCIL scenarios.

\section{Methodology}

%\subsection{Problem Statement}
\textbf{Problem Statement:} FSCIL considers a combination of few-shot learning, where supervision is extremely scarce, and CL, where models must update repeatedly under strict resource budgets. The training procedure is organized into a base session and a sequence of incremental sessions. In the base session, the model is trained on a large set of classes $\mathcal{C}_0$ with abundant labeled data. Each incremental session $t \in \{1, \ldots, T_N\}$ then introduces a small number of novel classes $\mathcal{C}_t$, with only a few $K$ labeled samples $\mathcal{D}_t = \{(x_i, y_i)\}_{i=1}^{K}$ %provided 
per class. %The training data for session $t$ is denoted by $\mathcal{D}_t = \{(x_i, y_i)\}_{i=1}^{K}$, and the class sets %$\mathcal{C}_t$ 
%are 
The label space is disjoint across sessions $\mathcal{C}_i \cap \mathcal{C}_j = \varnothing, \forall i \neq j$.
During each session, only data from that session is accessible, and models must adapt under deployment-level constraints with limited training time and restricted hardware resources while mitigating forgetting of past knowledge. 

\subsection{Selective Backpropagation}
\label{sec:sbp}

\begin{figure*}[t]
  \centering
  \includegraphics[width=0.80\textwidth]{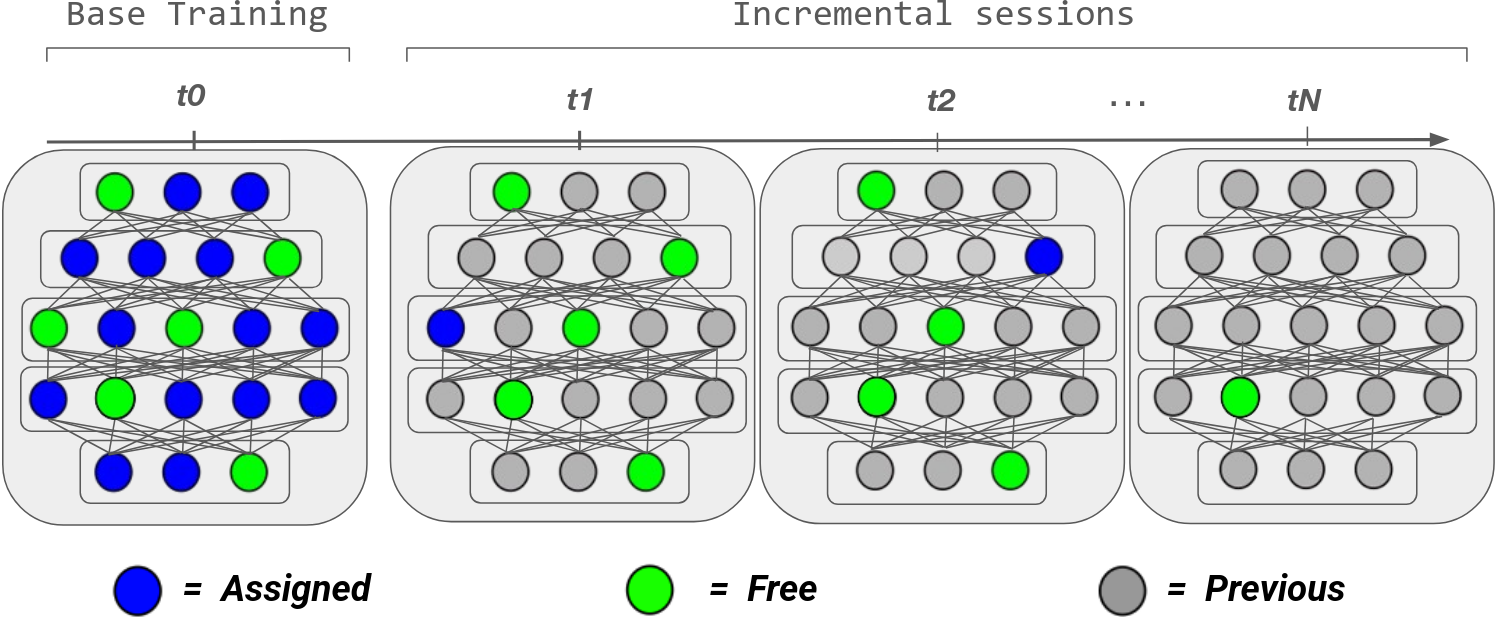}
  \caption{Selective Backpropagation (SBP) overview: parameter partitioning and masked gradient flow across incremental sessions.}
  \label{fig:sbp_overview}
\end{figure*}

% Motivated by these limitations, we propose Selective Backpropagation (SBP). Instead of tuning all parameters or none, SBP updates only a carefully chosen subset of the model’s parameters during incremental training. Unlike prior methods, SBP allows the model to acquire novel features while keeping most of the network unchanged. The computational and memory overhead of incremental updates is bounded and relatively small. By restricting gradient updates through a subset of parameters, SBP concentrates computation on regions of the network that retain unused representational capacity.

% In few-shot class-incremental learning (FSCIL), models must learn novel classes from extremely limited data while retaining performance on previously learned classes. Fine-tuning a shared backbone with new-task data leads to catastrophic forgetting, as gradients modify parameters critical for earlier tasks. Regularization-based methods alleviate this effect by softly constraining parameter updates, but they inevitably trade stability for plasticity and remain sensitive to hyperparameter choices.

% We adopt an alternative approach based on explicit parameter isolation. Rather than constraining interference, we eliminate it by enforcing hard separation between parameters used by different tasks. This motivates \emph{Selective Backpropagation} (SBP), a structured parameter budgeting mechanism that guarantees stability by design while preserving controlled plasticity.

In FSCIL, and CL in general, the standard practice of fine-tuning a shared backbone is inherently limited by the greedy nature of backpropagation (BP). Without constraints and experience replay, BP is computationally efficient but utilizes the entire network's plasticity to minimize the current loss, leading to forgetting of prior knowledge and overfitting on current data. %Partial updates to the network can potentially mitigate this problem, but usually at the expense of additional computational cost for finding important parameters in the network. Additionally, the model performance is also affected by reassigning low-utility parameters to new tasks that were previously trained for previous tasks.% This prematurely saturates representational capacity which should be preserved for future sessions. For incremental tasks with limited data, full network updates are not only destructive to prior knowledge but also largely redundant. 
To address this, we propose Selective Backpropagation (SBP), an alternative to standard BP that can partially update the network to mitigate forgetting without additional compute overload. The key idea of SBP is preemptive and random parameter assignment for each new task in FSCIL, freezing parameters assigned for past tasks, and random reinitialization of free parameters kept for future tasks. This allows the model to acquire novel features while keeping most of the network unchanged, without spending any extra compute on post-training analysis for finding important parameters for previous tasks. 
%To address this, we propose Selective Backpropagation (SBP). We argue that stability and long-term plasticity are best achieved through explicit parameter isolation rather than soft regularization. Methods that rely on selectively choosing critical parameters during training or through post-hoc analysis are computationally expensive. Instead, we avoid any compute overload through preemptive parameter assignment. SBP enforces a structured budget by partitioning the network into preserved parameters (for past tasks), active parameters (for the current task), and free parameters (reserved for the future). This allows the model to acquire novel features while keeping most of the network unchanged. Moreover, the computational and memory overhead of incremental updates is bounded and relatively small. By restricting gradient updates through a subset of parameters, SBP concentrates computation on regions of the network that retain unused representational capacity.

Figure~\ref{fig:sbp_overview} depicts an overview of the allocation mechanism in SBP during FSCIL. Let $\boldsymbol{\theta} = \{(\mathbf{W}_l, \mathbf{b}_l)\}_{l=1}^{L}$ denote the parameters of an $L$-layer neural network. At each incremental session, parameters in each layer $l$ are partitioned into three mutually exclusive subsets,
\begin{equation}
\boldsymbol{\theta}^{l}
=
\boldsymbol{\theta}^{l}_{\mathrm{assigned}}
\cup
\boldsymbol{\theta}^{l}_{\mathrm{prev}}
\cup
\boldsymbol{\theta}^{l}_{\mathrm{free}} .
\label{eq:sbp_partition}
\end{equation}

Where, $\boldsymbol{\theta}^{l}_{\mathrm{assigned}}$ denotes parameters optimized for the current task $t$, $\boldsymbol{\theta}^{l}_{\mathrm{prev}}$ denotes parameters learned in previous tasks and kept frozen at increment $t$, and $\boldsymbol{\theta}^{l}_{\mathrm{free}}$ denotes unallocated parameters reserved for future tasks. Note that parameter assignment is irreversible: at each session, the new assigned subset is sampled only from the remaining free parameters. Parameters assigned in previous sessions permanently enter $\theta_{\mathrm{prev}}$ and are never eligible for reassignment or further optimization. Thus, capacity follows the one-way transition $\theta_{\mathrm{free}}\rightarrow\theta_{\mathrm{assigned}} \rightarrow\theta_{\mathrm{prev}}$ across sessions. 

During training, gradients are allowed to flow only through $\boldsymbol{\theta}^{l}_{\mathrm{assigned}}$, whereas gradients corresponding to $\boldsymbol{\theta}^{l}_{\mathrm{prev}}$ and $\boldsymbol{\theta}^{l}_{\mathrm{free}}$ are explicitly blocked.
The blocking mechanism is implemented through a set of binary masks, 
$\mathbf{M}^{l}_{\text{assigned}}$, 
$\mathbf{M}^{l}_{\text{prev}}$, and 
$\mathbf{M}^{l}_{\text{free}}$, 
which correspond to each parameter partition and are enforced using a backward hook on the gradient computation. No parameter mask is applied to the forward pass at inference; the prototype classifier only restricts distance computation to the currently active feature dimensions, as described in Appendix A.1. Thus, making it suitable for all CL settings (e.g., TIL, CIL, FSCIL). 
The parameter update for layer $l$ at training step $k$ is therefore given by:
\begin{equation}
\boldsymbol{\theta}^{l}_{k+1}
\leftarrow
\boldsymbol{\theta}^{l}_{k}
-
\eta
\left(
\nabla_{\boldsymbol{\theta}^{l}} \mathcal{L}_t
\odot
\mathbf{M}^{l}_{\text{assigned}}
\right),
\label{eq:sbp_update}
\end{equation}
where $\eta$ denotes the learning rate, $\mathcal{L}_t$ is the task-specific loss, and $\odot$ denotes the Hadamard (element-wise) product. 
SBP assigns a subset of network capacity ($\theta_{\text{assigned}}^l$) at each session using an allocation ratio $\beta_t$. For FSCIL, we allocate a larger fraction in the base session $\beta_0$ than in later increments to leverage the abundant base data, and use a constant schedule $\beta_t=\rho$ for $t\ge 1$, where $\rho \in (0,1)$ (a hyperparameter). %\textcolor{blue}{We investigate the sensitivity of our method to the choice of the base session budget $\beta_0$ in Appendix~\ref{app:budgets}, demonstrating that SBP is robust to various capacity splits and that moderate base budgets (75\%--90\%) successfully recover most of the attainable performance.}
%a fixed allocation ratio $\rho \in (0,1)$ (a hyperparameter); in our implementation this corresponds to using a constant schedule $\beta_t=\rho$ for $t\ge 1$.
%For FSCIL, we allocate a larger fraction in the base session $\beta_0$ than in later increments to leverage the abundant base data. 
By restricting gradient updates to small assigned partitions (e.g. $\sim$1\% of the network per incremental session), SBP naturally acts as a strong structural regularizer that restricts the model from overparameterizing and overfitting on the sparse $K$-shot data. For more balanced CL settings, the allocation can instead be scaled with task size (e.g., number of classes).
% SBP enforces the partition via gradient masking in the backward path, the backbone forward computation remains unchanged.
%This formulation guarantees that update energy is exclusively directed toward the assigned parameters, while the norms of parameters in 
%$\boldsymbol{\theta}^{l}_{\text{prev}}$ and 
%$\boldsymbol{\theta}^{l}_{\text{free}}$ remain strictly invariant with respect to the task loss $\mathcal{L}_t$.
Freezing $\boldsymbol{\theta}^{l}_{\mathrm{prev}}$ during training prevents forgetting of past knowledge by preserving previously learned representations in the network. $\boldsymbol{\theta}^{l}_{\mathrm{prev}}$ also allows transfer of prior knowledge for efficient learning of new tasks, without catastrophic forgetting. %Importantly, freezing $\boldsymbol{\theta}^{l}_{\mathrm{free}}$ prevents incidental drift in unallocated parameters, ensuring that future tasks begin learning from a clean and unbiased initialization rather than from parameters influenced by earlier tasks. 
% Moreover, free parameters $\boldsymbol{\theta}^{l}_{\mathrm{free}}$ are crucially reinitialized in each training epoch to prevent the network from implicitly depending on unallocated capacity.

\subsection{Free Parameter Reinitialization}
%Moreover, u
Unlike prior masking methods that are only suitable for TIL, a key novelty of SBP that makes it suitable for CIL and FSCIL is the random reinitialization mechanism of the free parameters $\boldsymbol{\theta}^{l}_{\mathrm{free}}$ in each training epoch. %Unlike prior works, keeping free parameters $\boldsymbol{\theta}^{l}_{\mathrm{free}}$ and randomly reinitializing them in each training epoch 
This ensures that current tasks never become dependent on these parameters, and network performance for current tasks remains robust even when these parameters are updated for future tasks. A natural concern is whether free parameters, which are neither frozen nor explicitly masked at inference time, may inject noise into the learned representation. In SBP, free parameters are reinitialized at the beginning of every training epoch and receive no gradient updates. As a result, they are inherently non‑stationary during optimization: the model never observes a consistent mapping through the free subspace across updates. This prevents the network from encoding task‑relevant structure in these parameters and forces the optimizer to learn solutions that use previously allocated and currently assigned capacity while remaining invariant to the free partition. Consequently, although inference is deterministic and no forward masking is applied, the learned representation does not rely on free parameters in practice. %This behavior is empirically validated in Section~4.2, where replacing full‑space inference with a strictly masked partial‑distance classifier yields nearly identical performance. 
Appendix~\ref{sec:theory_reset} provides further theoretical reasoning for free parameter reinitialization, and Section~\ref{sec:ablation_study} and Appendix~\ref{sec:tsne_analysis} empirically demonstrate its effectiveness.% of random reinitialization of free parameters. %We provide further theoretical reasoning for free parameter initialization in Appendix~\ref{sec:theory_reset}.

\subsection{%Metric-Agnostic 
Prototype-based Classification}
Given the limited number of samples during incremental rounds in FSCIL, we initialize classifier weights for new classes using weight imprinting~\cite{qi2018lowshotlearningimprintedweights}, which provides a deterministic initialization by directly aligning weights with each class centroid, computed as a mean of the feature vectors of the training set in each increment. Prototypes are frozen after each increment and are not updated in later tasks. Frozen prototypes, however, can become misaligned with a changing feature representation in the backbone~\cite{de2021continual, gomez2024exemplar}. While SBP's parameter masking limits such changes, incremental optimization can still modify the unallocated portion of the representation, causing stored prototypes and current embeddings to differ along plastic dimensions. To reduce this mismatch, we use a partial-distance formulation that restricts prototype comparisons to feature dimensions allocated by SBP up to the current session. As new capacity is assigned, the active classification subspace expands accordingly, while dimensions reserved for future adaptation remain excluded. Additional details are provided in Appendix~\ref{appendix:pa_norm}.

During inference, we classify each sample using a nearest-class-mean classifier over the stored prototypes. Given a sample embedding \(f_\theta(x)\) and class prototype \(p_c\), we define the class logit as the negative distance between them:
\[
z_c(x) = -d\left(f_\theta(x), p_c\right),
\]
where \(d(\cdot,\cdot)\) denotes the partial-distance function described in Appendix~\ref{appendix:pa_norm}. The prediction probability is then obtained by applying a softmax over the logits of all observed classes:
\[
p(y=c\mid x)
=
\frac{\exp(z_c(x))}{\sum_{c'} \exp(z_{c'}(x))}.
\]

\section{Experiments}
\label{sec:experiments}
%\textbf{Datasets and metrics.} 
We first evaluate our method on three well-established benchmark datasets for FSCIL: CIFAR-100~\cite{krizhevsky2009learning}, MiniImageNet~\cite{vinyals2016matching} and the Caltech--UCSD Birds-200-2011 (CUB-200)~\cite{wah2011caltech} (in Appendix~\ref{app:cub}). %CIFAR-100 consists of 60{,}000 RGB images of size $32 \times 32$ across 100 classes, with 50{,}000 training images and 10{,}000 test images. MiniImageNet is a subset of ImageNet that contains 100 classes. Each class has 600 images, for a total of 60{,}000 images, all resized to $84 \times 84$. 
Dataset details are included in Appendix~\ref{app:experimental_setup}. For CIFAR-100 and MiniImageNet, we follow the standard FSCIL protocol established in~\cite{tao2020fewshotclassincrementallearning}. Specifically, the base session trains on 60 classes with full supervision. The remaining 40 classes are then learned incrementally across 8 sessions, each introducing 5 new classes under a 5-way 5-shot setting. Overall, the model learns all classes over 9 sequential sessions without revisiting old data. %As the evaluation metric, 
We report %average 
per-session mean-class accuracy %($A_{\text{inc}}$) 
and average incremental accuracy to measure performance, and per-session training time to measure computational efficiency. We also evaluate cross-domain transferability on MiniImageNet and CUB-200 datasets and long-horizon scalability on ImageNet-1K over 80 incremental sessions.%, with results reported in Appendix~\ref{app:vit_long_horizon}.

%We evaluate our method on the CIFAR-100, MiniImageNet benchmarks under the standard FSCIL protocol, where models are trained on 60 base classes and incrementally learn the remaining 40 classes over 8 sessions in a 5-way 5-shot setting. 
%We report %average 
%per session mean-class accuracy %($A_{\text{inc}}$) 
%and average incremental accuracy to measure performance, and per-session training time to measure computational efficiency. 

We compare against a broad set of recent FSCIL baselines (details in Appendix~\ref{app:experimental_setup}) on all datasets. All baselines are reproduced under the same experimental settings and with the ResNet-18 backbone as our method for a fair comparison. Results with additional backbone architectures for SBP are reported in Appendix~\ref{app:architectures}.
%On CIFAR-100, for methods we reproduced, we standardized the backbone to ResNet-18 and set the base session epoch count to 130, to match our experimental setting and avoid confounding due to architectural differences. We do the same for methods reproduced on MiniImageNet, but for 200 base session epochs. \textcolor{red}{Crucially, to maintain strict experimental integrity, we do not perform validation on the test set. We observed that several official implementations of prior methods implicitly tune base-session hyperparameters or select checkpoints using test-set feedback.} For SBP, we set an initial budget slice of $\beta = 0.85$, where $\beta$ controls the proportion of neurons in each layer selected from the total capacity and assigned to the current session. The remaining capacity is distributed across incremental sessions. 
We report accuracy as the %the 
mean over five independent runs using %fixed global 
random seeds for all experiments. %the experiments on CIFAR-100 and MiniImageNet. 
Complete details on implementation, hyperparameters, datasets, and baselines are provided in Appendix~\ref{app:experimental_setup}. %We also report results on the CUB-200-2011 dataset in Appendix~\ref{app:cub}. %Our code is available in the supplementary materials.

% \subsection{Experimental Setup}
% \textbf{Datasets and metrics.} We evaluate our method on two well-established benchmark datasets for FSCIL, CIFAR-100~\cite{Krizhevsky09learningmultiple} and MiniImageNet~\cite{vinyals2017matchingnetworksshotlearning}. CIFAR-100 consists of 60{,}000 RGB images of size $32 \times 32$ across 100 classes, with 50{,}000 training images and 10{,}000 test images. MiniImageNet is a subset of ImageNet that contains 100 classes. Each class has 600 images, for a total of 60{,}000 images, all resized to $84 \times 84$. For CIFAR-100 and miniImageNet, we follow the standard FSCIL protocol established in~\cite{tao2020fewshotclassincrementallearning}. Specifically, the base session trains on 60 classes with full supervision. The remaining 40 classes are then learned incrementally across 8 sessions, each introducing 5 new classes under a 5-way 5-shot setting. Overall, the model learns all classes over 9 sequential sessions without revisiting old data. As the evaluation metric, we report the %Average Incremental Accuracy 
% mean class accuracy per incremental session($A_{\text{inc}}$), and average accuracy over all increments. We also report the per-session total training runtimes to evaluate model efficiency. %to compare against SOTA methods.
% % Average Incremental Accuracy ($A_{\text{inc}}$) across all sessions and the Performance Drop (PD) ($A_0 - A_T$) to quantify catastrophic forgetting. 

\begin{table*}[!t]
\centering
\caption{CIFAR-100 FSCIL (5-way 5-shot) classification accuracy (\%) with ResNet-18 using the TOPIC~\cite{cheraghian2021semantic} class split. Results are over five runs with different random seeds. Best (per column) is in \textbf{bold}, second best is \underline{underlined}. For efficiency, lower is better.} %For methods where per-stage timing is unavailable, we report \texttt{-}.}
\footnotesize
\setlength{\tabcolsep}{3.0pt}
\renewcommand{\arraystretch}{1.15}
\resizebox{\textwidth}{!}{%
\begin{tabular}{lcccccccccccc}
\toprule
\textbf{Method} &
\multicolumn{9}{c}{\textbf{Sessions (Accuracy \%)}} & 
\multicolumn{3}{c}{\textbf{Efficiency}} \\
\cmidrule(lr){2-10}\cmidrule(lr){11-11}\cmidrule(lr){12-13}
 & \textbf{1} & \textbf{2} & \textbf{3} & \textbf{4} & \textbf{5} & \textbf{6} & \textbf{7} & \textbf{8} & \textbf{9} & \textbf{Avg} & \textbf{Time (m)} & \textbf{vs Ours} \\
\midrule
Entropy-Reg \cite{liu2022few}
& 74.37 & 71.31 & 68.93 & 66.61 & \underline{64.43} & \underline{62.30} & \underline{60.29} & \underline{58.40} & \underline{56.60}
& 64.80 & 486 & $34.71\times$ \\

ALICE \cite{peng2022few}
& 79.00 & 70.50 & 67.10 & 63.40 & 61.20 & 59.20 & 58.10 & 56.30 & 54.10
& 63.21 & 196 & $14.00\times$ \\

FACT \cite{zhou2022forward}
& 80.13 & 75.09 & 70.51 & 66.43 & 62.55 & 59.33 & 56.87 & 54.42 & 51.95
& 64.14 & 40 & $2.86\times$ \\

WaRP \cite{kim2023warping}
& 80.31 & \underline{75.86} & \underline{71.87} & \underline{67.58} & 64.39 & 61.34 & 59.15 & 57.10 & 54.74
& \underline{65.82} & \underline{20} & \underline{$1.43\times$} \\

NC-FSCIL \cite{yang2023neural}
& 79.57 & 59.23 & 61.03 & 61.76 & 58.79 & 55.28 & 53.74 & 51.96 & 49.43
& 58.98 & 85 & $6.07\times$ \\

OrCo \cite{ahmed2024orco}
& 79.83 & 72.79 & 69.71 & 65.09 & 61.16 & 57.59 & 54.39 & 51.82 & 49.38
& 62.42 & 28 & $2.00\times$ \\

Comp \cite{zou2024compositional}
& 79.17 & 69.11 & 63.61 & 59.01 & 55.22 & 52.09 & 50.01 & 48.27 & 45.67
& 58.02 & 89 & $6.36\times$ \\

Approx \cite{wang2024approximation}
& \underline{81.00} & 73.35 & 69.34 & 65.40 & 62.31 & 59.15 & 56.59 & 54.59 & 52.47
& 63.80 & 296 & $21.14\times$ \\

ADBS \cite{li2025adaptive}
& 80.53 & 74.34 & 69.03 & 64.43 & 60.40 & 56.85 & 53.69 & 50.86 & 48.32
& 62.05 & 27 & $1.93\times$ \\

SoftNet \cite{kang2024continual}
& 79.97 & 75.75 & 71.76 & 67.36 & 64.09 & 60.91 & 59.07 & 56.94 & 54.76
& 65.62 & 67 & $4.79\times$ \\

\midrule
No Replay
& 81.42 & 51.40 & 23.04 & 14.68 & 2.86 & 1.92 & 1.34 & 1.05 & 1.00
& 19.86 & 11 & $0.79\times$ \\

Partial Replay
& 81.05 & 72.83 & 66.76 & 61.19 & 56.91 & 52.20 & 49.88 & 46.61 & 44.99
& 59.16 & 11 & $0.79\times$ \\

Full Replay
& 81.42 & 76.46 & 71.47 & 66.56 & 62.06 & 58.85 & 55.59 & 53.13 & 50.50
& 64.00 & 16 & $1.14\times$ \\

Full Replay (Proj.) 
& 84.00
& 78.50
& 73.30
& 68.80
& 64.60
& 61.20
& 58.20
& 55.40
& 53.00
& 66.33
& 11
& $0.79\times$ \\

\midrule

\textbf{Ours}
& \textbf{82.70}
& \textbf{78.40}
& \textbf{74.40}
& \textbf{70.90}
& \textbf{67.50}
& \textbf{64.90}
& \textbf{62.00}
& \textbf{59.90}
& \textbf{57.90}
& \textbf{68.73}
& \textbf{14}
& \textbf{1.00}\bm{$\times$} \\

$\pm$ Std
& 0.20
& 0.40
& 0.40
& 0.40
& 0.40
& 0.50
& 0.40
& 0.30
& 0.30
& 0.37
& --
& -- \\

\bottomrule
\end{tabular}
}
\label{tab:fscil_sessions_total_runtime}
\end{table*}

\subsection{Experiments on CIFAR-100 and MiniImageNet}

Table~\ref{tab:fscil_sessions_total_runtime} reports the per-increment accuracy, average incremental accuracy, and total training time of FSCIL methods, including ours, on the CIFAR-100 dataset.
SBP attains 57.90\% accuracy (1.30\% higher than the second best) after the final session and 68.73\% average incremental accuracy (2.91\% higher than the second best), achieving state-of-the-art performance across all increments.

For efficiency, SBP trains in 14 minutes total across all sessions. Among the compared FSCIL methods, the next fastest is WaRP at 20 minutes, requiring 1.43$\times$ the training time of SBP. We note that the second-best method differs depending on the metric considered. For example, Entropy-Reg is the next best method in terms of final-session accuracy, but it requires 34.71$\times$ the training time of SBP. In contrast, WaRP is the second-best FSCIL method in terms of training time, but achieves 3.16\% lower final-session accuracy. 

To further demonstrate the effectiveness of our approach in balancing efficiency and performance, we also report results for fine-tuning (No Replay), Partial Replay (storing 5 exemplars per base and incremental class), and Full Replay baselines.
As expected, naive fine-tuning without replay suffers from catastrophic forgetting, setting a lower bound on accuracy. However, it provides a strong efficiency baseline, requiring only 11 minutes of total training time. Storing a limited memory buffer of 5 samples per class partially mitigates this degradation, achieving a final accuracy of 44.99\% and an average incremental accuracy of 59.16\%, while requiring the same 11 minutes of training. FSCIL methods generally mitigate forgetting more effectively than these simple baselines, but often at the expense of substantially greater training time. In contrast, our exemplar-free method outperforms Partial Replay by 12.91\% in final accuracy and 9.57\% in average incremental accuracy, while requiring only 14 minutes of total training time. This shows that SBP achieves a favorable operating regime, providing substantially stronger stability and plasticity with only a small increase in training cost over simple replay.

Additionally, while Full Replay is treated as a strong reference baseline because it retains all historical training data, our results in Table~\ref{tab:fscil_sessions_total_runtime} show that access to all previous samples does not necessarily yield the strongest performance in the FSCIL setting. Full Replay requires 16 minutes of total training time and achieves a final-session accuracy of 50.50\% and an average incremental accuracy of 64.00\%. We also evaluate \emph{Full Replay (Proj.)}, where the backbone is frozen after the base session and only a two-layer projection head is fine-tuned using full replay. This stronger baseline reaches 66.33\% average and 53.00\% final accuracy, but SBP still performs better without storing past data. Several FSCIL methods require substantially more training time while yielding only modest improvements over Full Replay, and some perform worse. In contrast, SBP outperforms Full Replay by 7.40\% in final-session accuracy and 4.73\% in average incremental accuracy, while also reducing total training time from 16 to 14 minutes. These results highlight the importance of evaluating FSCIL methods through an explicit stability--efficiency trade-off: if simple baselines already provide competitive performance at low computational cost, new FSCIL methods should justify their additional complexity through clear improvements in accuracy, efficiency, or both.

Table~\ref{tab:miniimagenet} shows the accuracy and training-time comparison of our method and FSCIL baselines on MiniImageNet.
Our method remains competitive on this more challenging benchmark and achieves strong accuracy numbers in most sessions. SBP achieves 54.10\% final-session accuracy and 65.26\% average incremental accuracy, remaining highly competitive compared to ALICE at 55.70\% final-session accuracy and OrCo at 64.65\% average incremental accuracy. While WaRP requires less training time (27 minutes) than SBP (30 minutes), it achieves approximately 10\% lower final and average incremental accuracy. Together with the CIFAR-100 results, these findings indicate that SBP provides a favorable stability--plasticity--efficiency trade-off. Finally, SBP maintains consistently strong performance across datasets and sessions, whereas the second-best method varies across benchmarks and metrics. This suggests that competing approaches may be more sensitive to the dataset or evaluation setting, while SBP provides more consistent performance across the evaluated FSCIL scenarios. Together, the CIFAR-100 and MiniImageNet results show that SBP provides a favorable accuracy--efficiency trade-off under the standard FSCIL protocol. However, these benchmarks contain only eight incremental sessions and draw base and novel classes from the same underlying distribution. They therefore provide limited evidence about whether a method remains plastic under distribution shift or stable over much longer learning sequences. We evaluate on these two settings next.

\begin{table*}[t]
\centering
\caption{MiniImageNet FSCIL (5-way 5-shot) classification accuracy (\%) and training efficiency with ResNet-18. Results are over five runs with different random seeds. Best (per column) is in \textbf{bold}, second best is \underline{underlined}. For efficiency, lower is better.}
\footnotesize
\setlength{\tabcolsep}{3.0pt}
\renewcommand{\arraystretch}{1.15}
\resizebox{\textwidth}{!}{%
\begin{tabular}{lcccccccccccc}
\toprule
\textbf{Method} &
\multicolumn{9}{c}{\textbf{Sessions (Accuracy \%)}} & 
\multicolumn{3}{c}{\textbf{Efficiency}} \\
\cmidrule(lr){2-10}\cmidrule(lr){11-11}\cmidrule(lr){12-13}
 & \textbf{1} & \textbf{2} & \textbf{3} & \textbf{4} & \textbf{5} & \textbf{6} & \textbf{7} & \textbf{8} & \textbf{9} & \textbf{Avg} & \textbf{Time (m)} & \textbf{vs Ours} \\
\midrule

Entropy-Reg \cite{liu2022few}
& 72.49 & 66.93 & 62.21 & 58.24 & 54.74 & 51.59 & 48.86 & 46.42 & 44.43
& 56.21 & 573 & $19.10\times$ \\

ALICE \cite{peng2022few}
& 80.60 & 70.60 & 67.40 & 64.50 & 62.50 & 60.00 & 57.80
& \underline{56.80} & \underline{55.70}
& 63.99 & 314 & $10.47\times$ \\

FACT \cite{zhou2022forward}
& 70.45 & 64.73 & 60.58 & 57.44 & 54.32 & 51.60 & 49.00 & 47.00 & 45.46
& 55.62 & 41 & $1.37\times$ \\

WaRP \cite{kim2023warping}
& 68.63 & 64.43 & 60.44 & 57.82 & 55.42 & 52.64 & 50.02 & 48.19 & 46.89
& 56.05 & \textbf{27} & \textbf{$0.90\times$} \\

NC-FSCIL \cite{yang2023neural}
& 76.45 & 68.12 & 63.50 & 59.28 & 57.10 & 54.22 & 51.80 & 49.65 & 47.90
& 58.67 & 883 & $29.43\times$ \\

OrCo \cite{ahmed2024orco}
& \textbf{82.30} & 72.60 & 68.94 & 65.57 & 62.70 & 60.11 & 58.08 & 56.58 & 54.96
& 64.65 & 157 & $5.23\times$ \\

Comp \cite{zou2024compositional}
& 77.38 & 67.31 & 62.61 & 58.69 & 56.17 & 53.09 & 50.43 & 48.09 & 46.70
& 57.83 & 290 & $9.67\times$ \\

Approx \cite{wang2024approximation}
& 38.27 & 24.69 & 22.93 & 21.83 & 20.68 & 19.64 & 18.64 & 18.14 & 17.58
& 22.49 & 91 & $3.03\times$ \\

ADBS \cite{li2025adaptive}
& \underline{81.65}
& \textbf{75.43}
& \textbf{71.43}
& \underline{67.99}
& \textbf{65.53}
& \underline{61.97}
& \textbf{59.04}
& \textbf{56.97}
& \textbf{55.74}
& \textbf{66.19}
& 191
& $6.37\times$ \\

SoftNet \cite{kang2024continual}
& 68.22 & 62.97 & 58.90 & 55.80 & 53.26 & 50.91 & 48.56 & 46.77 & 44.98
& 54.49 & 79 & $2.63\times$ \\

\midrule

\textbf{Ours}
& 78.10
& \underline{73.80}
& \underline{70.20}
& \textbf{68.50}
& \underline{65.40}
& \textbf{62.70}
& \underline{58.20}
& 56.30
& 54.10
& \underline{65.26}
& \underline{30}
& \underline{$1.00\times$} \\

$\pm$ Std
& 0.10
& 0.20
& 0.20
& 0.40
& 0.40
& 0.40
& 0.40
& 0.60
& 0.90
& 0.40
& --
& -- \\

\bottomrule
\end{tabular}
}
\label{tab:miniimagenet}
\end{table*}

\begin{table*}[t]
\centering
\caption{Cross-Domain FSCIL (MiniImageNet $\rightarrow$ CUB-200) classification accuracy (\%) %. The protocol is 
with an initial session of 60 base classes and 10 incremental sessions of 20-way 5-shot using a ResNet-18 backbone. Best (per column) is in \textbf{bold}, second best is \underline{underlined}.}
\small
\setlength{\tabcolsep}{2.8pt} 
\renewcommand{\arraystretch}{1.00}
\resizebox{\textwidth}{!}{%
\begin{tabular}{lccccccccccccc}
\toprule
\textbf{Method} &
\multicolumn{11}{c}{\textbf{Sessions (Accuracy \%)}} & \textbf{Avg (\%)} \\
\cmidrule(lr){2-12}\cmidrule(lr){13-13}
 & \textbf{1} & \textbf{2} & \textbf{3} & \textbf{4} & \textbf{5} & \textbf{6} & \textbf{7} & \textbf{8} & \textbf{9} & \textbf{10} & \textbf{11} &  \\
\midrule

ALICE~\cite{peng2022few} & 65.56 & 56.05 & 51.73 & 47.68 & 44.96 & 41.81 & 39.37 & 37.19 & 35.26 & 33.45 & 31.88 & 44.09 \\
FACT~\cite{zhou2022forward}
& \textbf{79.20} & 62.46 & 51.83 & 44.39 & 39.07 & 35.06 & 31.38 & 28.70 & 26.13 & 24.41 & 22.57 & 40.47 \\
WaRP \cite{kim2023warping} 
& 69.41 & \underline{64.13} & \underline{59.33} & \underline{54.81} & \underline{51.37} & \underline{48.23} & \underline{45.54} & \underline{43.06} & \underline{41.09} & \underline{38.39} & \underline{36.12} & \underline{50.13} \\
NC-FSCIL~\cite{yang2023neural}
& 64.35 & 54.27 & 50.17 & 44.15 & 39.47 & 35.64 & 32.46 & 29.82 & 27.37 & 25.47 & 23.81 & 38.82 \\
OrCo \cite{ahmed2024orco} 
& 58.05 & 43.95 & 39.89 & 34.38 & 30.35 & 29.62 & 28.33 & 26.50 & 24.25 & 23.24 & 22.07 & 32.78 \\
Comp \cite{zou2024compositional} 
& 59.97 & 55.48 & 50.98 & 47.56 & 44.37 & 41.52 & 39.05 & 36.74 & 34.73 & 32.88 & 31.19 & 43.13 \\
% Approx \cite{approx} 
% & 56.00 & 51.23 & 47.01 & 43.39 & 40.46 & 37.80 & 35.88 & 33.79 & 31.83 & 30.15 & 28.62 & 39.65 \\
% ADBS \cite{adbs} \\
SoftNet~\cite{kang2024continual}
& 67.63 & 60.98 & 54.94 & 50.46 & 46.03 & 42.74 & 39.90 & 37.25 & 34.88 & 32.79 & 31.08 & 45.33 \\
\midrule

\textbf{Ours}
& \underline{77.33} & \textbf{68.47} & \textbf{62.92} & \textbf{58.15} & \textbf{54.25} & \textbf{50.84} & \textbf{47.85} & \textbf{45.13} & \textbf{42.74} & \textbf{40.47} & \textbf{38.57} & \textbf{53.34} \\

\bottomrule
\end{tabular}%
}
\label{tab:cross_domain_results}
\end{table*}

\vspace{-0.9em}
\subsection{Cross-Domain FSCIL}
\label{sec:cross_domain_results}
~\cite{chen2019closer} showed that most few-shot learning methods fail in scenarios where base and novel classes are drawn from slightly different datasets. Therefore, %to further investigate the adaptability of our method, 
we adapt the MiniImageNet-CUB cross-domain scenario reported in~\cite{chen2019closer} to FSCIL to investigate our method and FSCIL baselines on this challenging setup. In this setting, 
%To evaluate robustness under distribution shift, we additionally consider a cross-domain FSCIL setting where 
the base session is trained on 60 classes on MiniImageNet, and incremental sessions introduce novel classes from CUB-200 in the form of 20-way 5-shot tasks. %We choose to compare against SoftNet which is another parameter masking method, and the two best performing FSCIL methods (FACT and NC-FSCIL) that freeze the backbone after base training. 
All methods use the ResNet-18 backbone trained from scratch, and are trained for 100 epochs for the base session, and 25 epochs for the incremental sessions. We further validate cross-domain robustness on MiniImageNet$\rightarrow$Caltech-101, with results in Appendix~\ref{app:cross_domain_caltech}. %(Table~\ref{tab:cross_domain_results}). We consider a few of the strongest performing baselines for this experiment. This benchmark is substantially harder than in-domain FSCIL as the model must retain the base-domain decision boundaries while rapidly adapting to fine-grained bird categories from a completely different domain. All methods were run with 100 epochs for the base session, and 25 incremental iterations for the incremental sessions.

Table~\ref{tab:cross_domain_results} reports the results of the cross-domain FSCIL experiment. SBP achieves the best accuracy for all incremental sessions, with a final accuracy of 38.57\% and an average accuracy of 53.34\%, %which are 2.45\% and 3.21\%, respectively, higher than the next best method (WaRP). 
outperforming the next-best method, WaRP, by 2.45\% and 3.21\%, respectively. We also observe a substantial degradation for FSCIL methods that freeze the backbone after the base session, as their fixed representations cannot adapt when incremental classes come from a distribution different from the base classes. This highlights a key limitation of many efficient FSCIL methods: they reduce computation by sacrificing adaptability, making them sensitive to domain shifts. In realistic CL settings, where future data may differ semantically or statistically from the base distribution, such rigidity is undesirable. SBP avoids this failure mode by maintaining controlled plasticity across sessions, allowing the backbone to adapt while limiting catastrophic forgetting. More broadly, these results illustrate that strong retention on a distribution-consistent FSCIL benchmark does not necessarily imply an effective stability--plasticity trade-off. Restricting representation changes can preserve past performance while also limiting the model's ability to learn shifted future classes.

\subsection{Long-Horizon FSCIL}
\label{sec:long_horizon}

Standard FSCIL benchmarks such as CIFAR-100 and MiniImageNet contain only eight incremental sessions, making it difficult to assess whether a method remains stable over substantially longer learning sequences. We therefore construct a long-horizon ImageNet-1K benchmark with 600 base classes followed by 400 novel classes introduced over 80 incremental sessions, each using a 5-way 5-shot protocol. All methods use a ResNet-18 backbone and are evaluated over all classes observed up to each session. %Full implementation details and additional experiments are provided in Appendix~\ref{app:vit_long_horizon}.

Table~\ref{tab:vit_long_horizon} shows that differences between methods become substantially more pronounced as the learning horizon increases. WaRP remains competitive during the early sequence and slightly outperforms SBP at Sessions~20 and~30, but its accuracy subsequently declines from 51.40\% at Session~30 to 12.90\% at Session~80. In contrast, SBP retains 39.08\% accuracy after all 80 incremental sessions. ADBS and OrCo similarly reach only 14.50\% and 18.40\% at the final session.

These results show that performance over short FSCIL sequences does not necessarily predict long-horizon behavior. In particular, methods that appear competitive over the first few increments can accumulate substantial degradation as the sequence grows. This reinforces the need to evaluate FSCIL methods not only on short, distribution-consistent benchmarks, but also under longer learning horizons that more directly stress the stability--plasticity trade-off.

\begin{table*}[!t]
\centering
\caption{Long-horizon FSCIL on ImageNet-1K. We report classification accuracy (\%) at selected sessions throughout the incremental sequence. Best (per column) is in \textbf{bold}, second best is \underline{underlined}.}
\footnotesize
\setlength{\tabcolsep}{3.5pt}
\renewcommand{\arraystretch}{1.15}
\resizebox{\textwidth}{!}{%
\begin{tabular}{lccccccccccc}
\toprule
\textbf{Method} &
\multicolumn{11}{c}{\textbf{Sessions (Accuracy \%)}} \\
\cmidrule(lr){2-12}
& \textbf{0} & \textbf{1} & \textbf{5} & \textbf{10}
& \textbf{20} & \textbf{30} & \textbf{40} & \textbf{50}
& \textbf{60} & \textbf{70} & \textbf{80} \\
\midrule

ADBS \cite{li2025adaptive}
& 61.50
& 57.10
& 47.20
& 39.20
& 27.40
& 22.20
& 18.80
& 17.20
& 16.00
& 14.90
& 14.50 \\

OrCo \cite{ahmed2024orco}
& 27.00
& 24.80
& 24.90
& 25.10
& 24.30
& 23.20
& 21.90
& 21.00
& 20.00
& 19.30
    & \underline{18.40} \\

WaRP \cite{kim2023warping}
& \underline{63.20}
& \underline{61.90}
& \underline{60.40}
& \underline{58.60}
& \textbf{54.90}
& \textbf{51.40}
& \underline{47.80}
& \underline{42.10}
& \underline{33.30}
& \underline{20.00}
& 12.90 \\

% \midrule
% ViT (SBP)
% & \textbf{73.56}
% & \textbf{70.19}
% & \textbf{65.03}
% & \textbf{63.28}
% & \textbf{58.85}
% & \textbf{55.78}
% & \textbf{53.08}
% & \textbf{50.77}
% & \textbf{48.93}
% & \textbf{46.85}
% & \textbf{45.07} \\

\midrule
\textbf{Ours}
& \textbf{72.45}
& \textbf{70.09}
& \textbf{64.99}
& \textbf{61.85}
& \underline{54.26}
& \underline{50.96}
& \textbf{47.93}
& \textbf{45.19}
& \textbf{43.01}
& \textbf{40.83}
& \textbf{39.08} \\

\bottomrule
\end{tabular}
}
\label{tab:vit_long_horizon}
\end{table*}

% The ViT variant provides the strongest long-horizon performance, retaining 45.07\% accuracy after 80 incremental sessions compared with 39.08\% for the ResNet-18 variant. Because the ViT is trained only on the 600 base classes and uses no external pretraining, this improvement reflects the ability of SBP to exploit a stronger backbone architecture rather than additional prior data. Overall, the results suggest that controlling interference becomes increasingly important as the number of incremental sessions grows.

\subsection{%Component 
Ablation Study}
\label{sec:ablation_study}
To analyze the contribution of each component in our method, we conduct an ablation study on CIFAR-100 using the same experimental setting as in Table~\ref{tab:fscil_sessions_total_runtime}. Results are reported in Table~\ref{tab:ablation}.
%Table~\ref{tab:ablation} quantifies the contribution of each component on CIFAR-100 under the exemplar-free FSCIL protocol. %Overall, performance is dominated by two factors: (i) stabilizing the representation across sessions, and (ii) restricting parameter updates to reduce interference.%
Our results show %that two main drivers of performance for our method are SBP, %'s mechanism of restricting parameter updates as well as 
%followed by %stabilizing outputs via 
%partial knowledge distillation (P-KD). 
that SBP is the primary driver of performance. %Removing the latter (\emph{w/o P-KD}) reduces average incremental accuracy from 68.69\% to 46.90\% and final accuracy from 56.07\% to 24.97\%. %Removing SBP (\emph{w/o SBP}) produces an even greater degradation in performance, dropping to 52.87\% average and 25.40\% final accuracy, consistent with severe interference under unrestricted updates. 
Removing SBP (\emph{w/o SBP}) leaves base-session performance intact (84.10\%) but causes immediate collapse once incremental learning begins, dropping to 13.90\% in Session~2 and 4.30\% by the final session. Average incremental accuracy falls from 68.73\% to 15.46\%, indicating severe interference from unrestricted parameter updates rather than insufficient base-session learning.

In contrast, \emph{SBP Only} (without a prototype-based classifier) retains most of the full model's performance, achieving 63.43\% average incremental accuracy and 48.85\% final-session accuracy, only 5.30 and 9.05 percentage points below the full model. Its gradual degradation indicates that selective parameter isolation is the main contributor to continual retention, while the remaining components mainly improve adaptation. Free-neuron re-initialization is also essential. Without it (\emph{w/o Free-Init}), accuracy falls from 82.57\% in the base session to 1.88\% in Session~2 and 1.56\% in the final session, yielding only 10.62\% average accuracy. This suggests that unreinitialized free parameters can become implicitly co-adapted to the current session, so reassigning them later disrupts previously learned features. We further analyze this failure mode with t-SNE visualizations in Appendix~\ref{sec:tsne_analysis}.

In line with prior few-shot learning research, prototype imprinting is a useful complementary component for FSCIL. Removing imprinting (\emph{w/o Imprinting}) reduces average incremental accuracy from 68.73\% to 62.63\% and final-session accuracy from 57.90\% to 48.24\%. Unlike the collapses observed when SBP or free-neuron re-initialization is removed, performance without imprinting declines gradually across sessions. This suggests that imprinting does not primarily prevent catastrophic forgetting, %, but instead improves %how the model adapts to new classes. This is particularly 
but it is important in FSCIL, where each novel class is represented by only five training examples and reliable initialization of its classifier is difficult to obtain through gradient-based optimization alone.
% Finally, the choice of distance metric has a relatively minor effect. Replacing Mahalanobis distance with Euclidean distance (\emph{w/o Mahalanobis}) reduces the average accuracy only slightly, from 68.52\% to 67.78\%, and the final accuracy from 58.52\% to 56.78\%. %Similarly, the partial-distance variant (\textit{P-Dist}), which removes explicit drift compensation and restricts comparisons to the active subspace, remains close to the full model, achieving 67.84\% average accuracy and 56.32\% final accuracy. This suggests that prototype drift can be mitigated either through explicit drift compensation or through subspace-consistent distance computation enabled by SBP's parameter-isolation structure. 
Additional ablations over backbone architectures and base-session budgets are provided in Appendices~\ref{app:architectures} and~\ref{app:budgets}.

\begin{table*}[t]
\centering
\small
\setlength{\tabcolsep}{3.8pt}
\renewcommand{\arraystretch}{1.15}
\caption{Ablation study on CIFAR-100 FSCIL, 5-way 5-shot task.
We report per-session classification accuracy (\%), average incremental accuracy, and the change in average accuracy relative to the full model (\(\Delta\)). Best (per column) is in \textbf{bold}, second best is \underline{underlined}.}
\label{tab:ablation}
\begin{tabular}{lccccccccc|cc}
\toprule
& \multicolumn{9}{c|}{Session Accuracy (\%)} & \multicolumn{2}{c}{Metrics} \\
\cmidrule(lr){2-10}\cmidrule(lr){11-12}
Ablation & 1 & 2 & 3 & 4 & 5 & 6 & 7 & 8 & 9 & Avg (\%) & $\Delta$ \\
\midrule

\textbf{Full}
& 82.70
& \textbf{78.40}
& \textbf{74.40}
& \textbf{70.90}
& \textbf{67.50}
& \textbf{64.90}
& \textbf{62.00}
& \textbf{59.90}
& \textbf{57.90}
& \textbf{68.73}
& --- \\

\midrule

SBP Only
& \underline{82.85}
& \underline{76.25}
& \underline{70.81}
& \underline{66.20}
& \underline{61.90}
& \underline{58.20}
& \underline{54.36}
& \underline{51.47}
& \underline{48.85}
& \underline{63.43}
& $-$5.30 \\

w/o Imprinting
& 82.70
& 75.75
& 69.99
& 65.01
& 60.70
& 57.18
& 53.76
& 50.77
& 48.24
& 62.63
& $-$6.10 \\

w/o SBP
& \textbf{84.10}
& 13.90
& 10.00
& 5.60
& 7.00
& 6.20
& 4.00
& 4.00
& 4.30
& 15.46
& $-$53.27 \\

w/o Free-Init
& 82.60
& 1.88
& 1.66
& 1.69
& 1.85
& 1.36
& 1.56
& 1.43
& 1.56
& 10.62
& $-$58.11 \\

\bottomrule
\end{tabular}%
\end{table*}

% \vspace{-8em}

\begin{comment}
We also notice that the distance metric choice has very little effect on overall performance. Replacing Mahalanobis with Euclidean distance (\emph{w/o Mahalanobis}) causes a slight drop in average accuracy (0.74\%) %but yields a slight increase in final accuracy (0.71\%)
and final accuracy (1.74\%), suggesting that it is not a primary forgetting mitigation mechanism. 
Finally, using partial distance (\emph{P-Dist}) (Appendix~\ref{appendix:pa_norm}), and removing drift compensation yields %almost the same 
similar performance as the full method, with a drop of 2.2\% and 0.68\% in final and average incremental accuracy, respectively. %\textcolor{red}{with only a slight drop in final accuracy (0.85\%) and a marginal increase in final accuracy (0.25\%).}
These results indicate that drift issues can be largely mitigated by enforcing subspace-consistent comparisons rather than requiring explicit drift correction, thanks to the parameter isolation mechanism of SBP. Further ablations using different backbones and base session parameter budgets are in Appendix~\ref{app:architectures} and~\ref{app:budgets}.
\end{comment}

\section{Related Work}
\label{sec:related_work}
Few-Shot Class-Incremental Learning (FSCIL) lies at the intersection of few-shot learning and class-incremental learning (CIL), and inherits challenges from both. In CIL, models must learn a growing set of classes without access to task identity at test time, making them particularly vulnerable to catastrophic forgetting~\cite{rebuffi2017icarl}. FSCIL further amplifies this difficulty by introducing only a handful of labeled examples per incremental class~\cite{tao2020fewshotclassincrementallearning}, which increases both overfitting on new classes and interference with previously learned ones~\cite{zhao2021mgsvf}. As a result, FSCIL methods must carefully balance stability, plasticity, and computational efficiency under strict data and memory constraints.

A broad class of FSCIL methods mitigates forgetting using replay, meta-learning, or heavily structured objectives. Meta-learning approaches aim to learn update rules or initializations that facilitate rapid adaptation to new classes~\cite{chi2022metafscil,zhou2022forward}, while others improve base–incremental compatibility via pseudo-features, or data-free replay mechanisms~\cite{cheraghian2021semantic,liu2022few}. Related work also emphasizes prototype separability and alignment, using prototype evolution, feature regularization, or geometry-aware losses to control representation drift~\cite{tao2020fewshotclassincrementallearning,hersche2022constrained, zhang2021few, zhu2021self}. While these approaches achieve strong accuracy in controlled benchmarks, they often require additional training stages, careful tuning of multiple loss terms, or replay-style refinement. As a result, their computational cost scales poorly, and performance can degrade when assumptions about task distributions or prototype structure are violated. %These methods therefore, tend to trade efficiency or robustness for stability, particularly in long or resource-constrained incremental streams.
To improve efficiency, several FSCIL methods freeze most or all of the backbone after the base session and restrict learning to lightweight classifier or prototype updates~ \cite{zhou2022forward, yang2023neural}. This strategy substantially reduces training cost and limits forgetting in the few-shot regime. However, it also implicitly assumes that features learned during base training remain sufficient for all future tasks, an assumption that breaks down in cross-domain FSCIL, resulting in severe performance degradation.

A promising line of work addresses forgetting by explicitly allocating network capacity to tasks through subnetworks, masking, or parameter isolation~\cite{mallya2018packnet, serra2018overcoming}. Early methods demonstrate strong performance in the task-incremental learning (TIL) setting, where task identity is available at test time, but they are unsuitable for FSCIL. More recent work adapts these ideas to FSCIL, for example, via sparse or ``winning ticket'' subnetworks~\cite{mazumder2021few,kang2024continual}. However, %most masking-based approaches either rely on task identity, 
these methods require learning importance masks through additional optimization, or incur nontrivial overhead due to per-task search, pruning, or retraining. Moreover, aggressive isolation can fragment capacity and limit forward transfer to later tasks. These limitations motivate FSCIL-compatible capacity allocation strategies that control interference deterministically, without task inference or expensive mask learning. 

In this direction, several recent works have explored using prompting mechanisms on large pre-trained foundation models (e.g., Vision Transformers) for FSCIL~\cite{li2025prompt,wang2024approximation}. These approaches depart from the strict, learning-from-scratch, FSCIL setting studied in this paper. They rely heavily on massive external pre-training datasets, making it difficult to disentangle incremental learning from prior knowledge embedded in the model. This dependence introduces leakage from pre-training and obscures the model’s true ability to acquire, retain, and adapt knowledge incrementally. Furthermore, because such models are typically frozen or only weakly adapted, they often exhibit limited robustness under domain shift and do not support genuinely continual representation learning. As a result, prompting-based methods are best viewed as a complementary direction rather than a solution to exemplar-free FSCIL under realistic deployment constraints.

\section{Conclusion}
\label{sec:conclusion}
This paper introduced \textit{Selective Backpropagation} (SBP), an efficient parameter-budgeting framework for exemplar-free few-shot class-incremental learning. SBP addresses a central tension in FSCIL: standard fine-tuning is computationally efficient and adaptive but prone to catastrophic forgetting, whereas many strong FSCIL methods preserve stability through costly replay, rigid backbone freezing, or limited plasticity. By preemptively allocating capacity, freezing parameters associated with past knowledge, and reinitializing future capacity at each epoch, SBP preserves stability while avoiding implicit dependence on unused parameters. When combined with %partial knowledge distillation and 
prototype-based classification, SBP achieves state-of-the-art performance across multiple FSCIL benchmarks while substantially reducing training cost relative to competing methods. Across standard benchmarks, cross-domain shifts, and an 80-session ImageNet-1K stream, our results show that strong performance on short, distribution-consistent FSCIL benchmarks alone is insufficient to characterize robust behavior over time. Methods that effectively preserve a fixed representation can remain competitive when future classes resemble the base distribution, yet degrade when actual adaptation or long-term interference becomes important. These findings motivate evaluating FSCIL together alongside stability, plasticity, and computational efficiency, with explicit evaluations for distribution shift and learning horizon. SBP performs strongly across these regimes, suggesting that controlled representational plasticity is an important component of practical FSCIL systems. 

\bibliography{iclr2027_conference}
\bibliographystyle{iclr2027_conference}

\appendix
\appendix

\section{Experimental Setup}
\label{app:experimental_setup}

\textbf{Implementation Details.} We utilize a ResNet-18~\cite{he2016deep} backbone as the feature extractor $f_\theta$ for all models, including ours. 
% For CIFAR-100, we use the standard low-resolution variant by removing the initial max-pooling layer.
% For MiniImageNet, we retain the $3 \times 3$ max-pooling layer with stride 2 after the initial convolution to accommodate the $84 \times 84$ spatial resolution. 
We use an SGD optimizer with momentum $0.9$ for all training sessions. During the base session, we train for 130 epochs on CIFAR-100 and 200 epochs on MiniImageNet. The learning rate is initialized to $\eta = 0.1$ and decayed using a cosine annealing schedule. After base training, similar to \cite{kang2024continual}, all batch normalization statistics are frozen to prevent representation drift. 
% caused by high-variance statistics from few-shot data. 
During incremental sessions, we fine-tune only the assigned parameters for 40 iterations on CIFAR-100 and 120 iterations on MiniImageNet using a reduced learning rate of $\eta = 0.0025$. We set an initial budget slice of $\beta = 0.85$, where $\beta$ controls the proportion of parameters in each layer selected from the total capacity and assigned to the current session. Specifically, this means $85\%$ of the parameters in each layer are assigned to the base classes. The remaining $15\%$ capacity is distributed across incremental sessions, allocating approximately $\sim 1.8\%$ of the parameters per layer to each new task. However, SBP does not rely on this knowledge, and remains robust against a wide range of parameter budgets (Appendix~\ref{app:budgets}). %in  \textcolor{blue}{In cases where the total number of incremental tasks $T_N$ is known beforehand, this incremental allocation can be computed precisely as $(1 - \beta_0) / T_N$, ensuring that the network capacity is exactly exhausted by the final session.} 
% The %initial 
% distillation weight is set to $\lambda = 7.0$, with a %starting 
% distillation temperature of $T = 5.0$ and an exponential decay factor of $\gamma = 0.99$. We use the Mahalanobis distance for our method in all experiments, % with diagonal entires of the covariance matrix only. 
% %We use 
% with a shared covariance matrix estimated from base-session features %with diagonal regularization 
% (diagonal entries only) and fixed for all subsequent incremental sessions. 
All experiments on CIFAR-100 and MiniImageNet are run on a single NVIDIA RTX A4500 GPU. The ablation study was conducted on an NVIDIA A100 Tensor Core GPU, utilizing a 20GB Multi-Instance GPU (MIG) partition. %Our code is available in the supplementary material. 
We report the mean and standard deviation over five independent runs using fixed global random seeds for CIFAR-100, MiniImageNet, and the CUB-200-2011 dataset experiments. The global random seed also determines the random SBP parameter allocation. Consequently, each independent run uses a different realization of the session-wise allocation masks, so the reported run-to-run variation also captures sensitivity to random parameter assignment. %We will publicly release our code after the paper is accepted.

\textbf{Datasets.} We evaluate our method on three well-established benchmark datasets for FSCIL: CIFAR-100~\cite{krizhevsky2009learning}, MiniImageNet~\cite{vinyals2016matching}, and the Caltech--UCSD Birds-200-2011 (CUB-200)~\cite{wah2011caltech}. CIFAR-100 consists of 60{,}000 RGB images of size $32 \times 32$ across 100 classes, with 50{,}000 training images and 10{,}000 test images. MiniImageNet is a subset of ImageNet that contains 100 classes. Each class has 600 images, for a total of 60{,}000 images, all resized to $84 \times 84$. CUB-200 consists of 200 bird species and 11,788 images. Following standard practice~\cite{kang2024continual}, all images are resized to $224\times224$. For CIFAR-100 and MiniImageNet, we follow the standard FSCIL protocol established in~\cite{tao2020fewshotclassincrementallearning}. Specifically, the base session trains a ResNet-18 model from scratch on 60 classes with full supervision. The remaining 40 classes are then learned incrementally across 8 sessions, each introducing 5 new classes under a 5-way 5-shot setting. Overall, the model learns all classes over 9 sequential sessions without revisiting old data. For CUB-200, a ResNet-18 backbone pre-trained on ImageNet-1K is finetuned for 60 epochs on the first 100 classes using all available samples, and the remaining 100 classes are introduced over 10 incremental sessions trained for 120 incremental iterations, each containing 10 novel classes under a 10-way 5-shot setting.

\textbf{Baselines.} We compare %our SBP 
our method against a strong set of representative FSCIL baselines. On CIFAR-100 and MiniImageNet, we evaluate methods such as Entropy-Reg~\cite{liu2022few}, ALICE~\cite{peng2022few}, FACT~\cite{zhou2022forward}, NC-FSCIL \cite{yang2023neural}, OrCo \cite{ahmed2024orco}, Comp \cite{zou2024compositional}, Approx \cite{wang2024approximation}, ADBS \cite{li2025adaptive}, SoftNet~\cite{kang2024continual}, and WaRP~\cite{kim2023warping}. We reproduced and evaluated all baselines to ensure a fair comparison. %. "We reproduced several methods—specifically 
% including Entropy-Reg, C-FSCIL, SoftNet, ALICE, WARP, FACT, and NC-FSCIL.%—to collect granular per-stage timing data
% . For these re-evaluated baselines, we report both the total training duration and the relative speedup compared to our method. For methods without timing, we include reported accuracies from prior work (cRT~\cite{shi2021overcomingcatastrophicforgettingincremental}, MetaFSCIL~\cite{Chi_2022_CVPR}, TOPIC~\cite{cheraghian2021semanticawareknowledgedistillationfewshot}, CEC~\cite{zhang2021fewshotincrementallearningcontinually}, F2M~\cite{shi2021overcomingcatastrophicforgettingincremental}, FSLL~\cite{mazumder2021few}) without re-measuring runtimes. 
On CUB-200-2011, we compare against some of the previously mentioned baselines including ALICE~\cite{peng2022few}, FACT~\cite{zhou2022forward}, WaRP~\cite{kim2023warping}, OrCo \cite{ahmed2024orco}, Comp \cite{zou2024compositional}, and NC-FSCIL~\cite{yang2023neural}. To ensure a fair comparison, we evaluated all models on the same settings and backbone used for SBP. Additionally, we fixed a few errors, such as the use of the test set as the validation set, for some methods to ensure a fair comparison. %we reproduced the results for SoftNet and NC-FSCIL on Mini-ImageNet under our own experimental settings, selecting them due to their superior reported performance on this dataset and in our experiment on CIFAR-100. We also note that reproducing some baselines from the released code required non-trivial effort due to dependency/version incompatibilities and implementation issues. In particular, NC-FSCIL and SoftNet required substantial debugging to both reproduce their own results and run under our evaluation protocol for a fair comparison. We acknowledge that SoftNet reports improved performance on CIFAR-100 (56.00\% final session accuracy on CIFAR-100) when utilizing a Fourier Subneural Operator (FSO). However, the source code for the FSO module has not been released, and our requests to the authors for the implementation were unsuccessful. As we cannot independently verify these results, we exclude this specific variant from our comparison.

\subsection{Partial Distance Classification}
\label{appendix:pa_norm}

Conventional distance-based classifiers operate over the full feature
embedding space $\mathbb{R}^D$, causing prototype comparisons to include
dimensions associated with capacity that has not yet been allocated for
learning. Under SBP, these dimensions may contain unstable activations
because their corresponding parameters are reserved for future
incremental sessions. We therefore restrict prototype comparisons to the
feature dimensions that have been allocated up to the current session.
As SBP progressively assigns new capacity, the active classification
subspace grows accordingly.

\textbf{Active Subspace Projection.}
For the convolutional backbones used in our standard experiments, the
output channels of the final convolution correspond directly to feature
dimensions after global pooling. Let
$\mathbf{M}_{\mathrm{assigned}}^{(j)}$ denote the feature dimensions
allocated at session $j$. At session $t$, we define the active feature
mask as
\[
\mathbf{M}_{\mathrm{active}}^{(t)}
=
\bigvee_{j=0}^{t}
\mathbf{M}_{\mathrm{assigned}}^{(j)},
\]
where $\lor$ denotes the element-wise logical OR. The corresponding
projection is
\[
\Phi_t(\mathbf{v})
=
\mathbf{v}\odot\mathbf{M}_{\mathrm{active}}^{(t)}.
\]
Thus, dimensions reserved for future sessions are excluded from
classification, while newly allocated dimensions become available to
the classifier as incremental learning proceeds.

\paragraph{Prototype Imprinting and Inference.}
For each new class $c$, we use standard weight imprinting and store its
prototype as the centroid of the full support features at the session
$t_c$ in which the class is introduced:
\[
p_c =
\frac{1}{|D_c|}
\sum_{x\in D_c} f_{\theta_{t_c}}(x).
\]
The prototype is then frozen. At inference during session $t$, distances
are computed over the feature dimensions allocated up to the current
session:
\[
d_c^{(t)}(x)
=
d\!\left(
M_{\mathrm{active}}^{(t)} \odot f_{\theta_t}(x),
M_{\mathrm{active}}^{(t)} \odot p_c
\right).
\]

This projection excludes dimensions that remain unallocated at session $t$, preventing currently free capacity from directly affecting classification. %Storing prototypes in the full feature space is important in this setting: when the active mask expands, earlier classes retain their original feature values in the newly activated dimensions rather than introducing artificial zeros into the distance computation.

\begin{table*}[t]
\centering
\caption{CIFAR-100 FSCIL (5-way 5-shot) per-session runtime (minutes) with ResNet-18 using the TOPIC class split. % All experiments were ran on a single Nvidia RTX A4500 GPU. 
%Best (lowest, per column) is in \textbf{bold}, second best is \underline{underlined}. %$^*$Originally used a different backbone, results shown are with ResNet-18 for fair comparison. 
The ``Backbone Update'' column indicates whether the backbone is tuned during incremental sessions or frozen after base training. Session 1 reports the base session (60 classes) runtime. Lower runtime is better; best and second-best values in each runtime column are shown in \textbf{bold} and \underline{underlined}, respectively. Results are averaged over five runs with different random seeds.}
\small
\setlength{\tabcolsep}{4.2pt}
\renewcommand{\arraystretch}{1.15}
\begin{tabular}{ll ccccccccc c}
\toprule
\textbf{Method} &
\textbf{Backbone} &
\multicolumn{9}{c}{\textbf{Sessions (time in minutes)}} &
\textbf{Total} \\
\cmidrule(lr){3-11}
 & & \textbf{1} & \textbf{2} & \textbf{3} & \textbf{4} & \textbf{5} & \textbf{6} & \textbf{7} & \textbf{8} & \textbf{9}
 & \textbf{(min)} \\
\midrule
Entropy-Reg \cite{liu2022few}
& Tuned & 56 & 52 & 52 & 53 & 53 & 54 & 55 & 55 & 56 & 486 \\
ALICE \cite{peng2022few}
& Frozen & 188 & 1 & 1 & 1 & 1 & 1 & 1 & 1 & 1 & 196 \\
FACT \cite{zhou2022forward}
& Frozen & 39 & \underline{0.03} & \underline{0.03} & \underline{0.03} & \underline{0.03} & \underline{0.03} & \underline{0.03} & \underline{0.03} & \underline{0.03} & 40 \\
WaRP \cite{kim2023warping}
& Tuned & \underline{19} & 0.08 & 0.05 & 0.05 & 0.05 & 0.05 & 0.05 & 0.05 & 0.06 & \underline{20} \\
NC-FSCIL \cite{yang2023neural}
& Frozen & 84 & \underline{0.03} & 0.04 & 0.06 & 0.06 & 0.07 & 0.08 & 0.08 & 0.08 & 85 \\
OrCo \cite{ahmed2024orco} 
& Frozen & 27 & 0.13 & 0.13 & 0.13 & 0.13 & 0.13 & 0.13 & 0.13 & 0.13 & 28 \\
Comp \cite{zou2024compositional} 
& Frozen & 88 & \textbf{0.01} & \textbf{0.01} & \textbf{0.01} & \textbf{0.01} & \textbf{0.01} & \textbf{0.01} & \textbf{0.01} & \textbf{0.01} & 89 \\
Approx \cite{wang2024approximation} 
& Frozen & 295 & \textbf{0.01} & \textbf{0.01} & \textbf{0.01} & \textbf{0.01} & \textbf{0.01} & \textbf{0.01} & \textbf{0.01} & \textbf{0.01} & 296 \\
ADBS \cite{li2025adaptive} & Frozen & 27 & \textbf{0.01} & \textbf{0.01} & \textbf{0.01} & \textbf{0.01} & \textbf{0.01} & \textbf{0.01} & \textbf{0.01} & \textbf{0.01} & 27 \\
SoftNet \cite{kang2024continual}
& Tuned & 66 & 0.06 & 0.06 & 0.06 & 0.06 & 0.06 & 0.06 & 0.06 & 0.06 & 67 \\
% C-FSCIL \cite{hersche2022constrainedfewshotclassincrementallearning}
% & Frozen & 85 & \textbf{0.01} & \textbf{0.01} & \textbf{0.01} & \textbf{0.01} & \textbf{0.01} & \textbf{0.01} & \textbf{0.01} & \textbf{0.01} & 86 \\
% LIMIT \cite{10.1109/TPAMI.2022.3200865}
% & Frozen & 34 & \underline{0.03} & \underline{0.03} & \underline{0.03} & \underline{0.03} & \underline{0.03} & \underline{0.03} & \underline{0.03} & \underline{0.03} & 35 \\
\midrule
\textbf{Ours}
& Tuned & \textbf{14} & 0.05 & 0.06 & 0.06 & 0.06 & 0.06 & 0.05 & 0.04 & 0.05 & \textbf{14} \\
\bottomrule
\end{tabular}
\label{tab:fscil_session_times}
\end{table*}

\section{Incremental Training Time Comparison with SOTA}
\label{app:timing}

We provide a breakdown of the training time per session for our method compared to current state-of-the-art (SOTA) methods. As aforementioned, the practicality of FSCIL is often limited by the computational cost of training in resource-constrained environments. 
Table~\ref{tab:fscil_session_times} reports the per-session runtimes on CIFAR-100 (5-way 5-shot) over nine sessions. For a fair comparison, all methods used the same backbone (ResNet-18), and were run on the same NVIDIA A4500 GPU. Our method achieves the fastest base-session training, completing Session~1 in approximately 14 minutes. This is $1.36\times$ faster than the closest competitor, WaRP, and almost %more than 
$5\times$ faster %than ALICE
than SoftNet.
% , which is the next best method in terms of accuracy. 
In contrast, several competing approaches require over one hour of base training, with ALICE and Approx exceeding 100 minutes. During incremental sessions (Sessions~2--9), our method consistently maintains low latency, requiring approximately 0.04--0.06 minutes per session. Methods that freeze the backbone after the base session (FACT, NC-FSCIL), generally achieve slightly better per-session training time (0.03 minutes) in most sessions. However, note that this comes at the cost of lower accuracy in incremental sessions, particularly in cases when the incremental sessions contain data that might be different from data used in the base session. Additionally, 
%While methods like LIMIT and FACT \cite{zhou2022forwardcompatiblefewshotclassincremental} report slightly lower per-session runtimes (0.03 minutes), 
these methods often require substantially longer base training to reach peak performance. SoftNet achieves similar per-session training time as our method after the base session, as both methods update only a small number of parameters in the backbone. Other methods, such as ALICE and WARP, %further 
incur significantly larger incremental overheads, ALICE incurs approximately 1 minute per incremental session, whereas WaRP requires approximately 0.05–0.08 minutes, on top of their high base session training times.
% These results indicate that our method not only surpasses baselines in accuracy but also achieves a superior balance for base-session training time and incremental update cost.
These results indicate that our method achieves a favorable balance between base-session training time and incremental update cost while achieving the best accuracy compared to SOTA baselines. %while remaining competitive with SOTA baselines.

\section{Results on CUB-200-2011}
\label{app:cub}

% As shown in Table~\ref{tab:new_results_11_sessions}, w
We report additional results on the Caltech--UCSD Birds-200-2011 (CUB-200) dataset, comprising 200 bird species and 11,788 images. %Following standard practice~\cite{softnet}, all images are resized to $224\times224$, and a ResNet-18 backbone pre-trained on ImageNet-1K is employed. \textcolor{red}{The base session is trained for 60 epochs on the first 100 classes using all available samples, and the remaining 100 classes are introduced over 10 incremental sessions trained for 120 incremental iterations, each containing 10 novel classes under a 10-way 5-shot setting. All methods were replicated using our experimental settings.}
All methods follow the same training settings and the backbone as reported in Appendix~\ref{app:experimental_setup}. Table~\ref{tab:new_results_11_sessions} reports the results on the CUB-200-2011 benchmark under the 10-way 5-shot FSCIL setting. SBP achieves the best accuracy across most sessions, with a final-session accuracy of 62.51\% and an average accuracy of 71.28\%. This corresponds to a 3.40\% improvement in final accuracy over the strongest competing method, Comp, and a 5.66\% improvement in average accuracy over Comp. These results further confirm that SBP generalizes beyond CIFAR-100 and MiniImageNet, providing robust performance across diverse FSCIL settings.

\begin{table*}[t]
\centering
\caption{CUB-200-2011 FSCIL (10-way 5-shot) classification accuracy (\%) with ResNet-18. Best (per column) is in \textbf{bold}, second best is \underline{underlined}. Results are averaged over five runs with different random seeds.%Avg is the mean over sessions 1--11. %Performance of other methods not re-measured.
}
\small
\setlength{\tabcolsep}{2.8pt} 
\renewcommand{\arraystretch}{1.15}
\resizebox{\textwidth}{!}{%
\begin{tabular}{lcccccccccccc}
\toprule
\textbf{Method} &
\multicolumn{11}{c}{\textbf{Sessions (Accuracy \%)}} &
\textbf{Avg (\%)} \\
\cmidrule(lr){2-12}\cmidrule(lr){13-13}
 & \textbf{1} & \textbf{2} & \textbf{3} & \textbf{4} & \textbf{5} & \textbf{6} & \textbf{7} & \textbf{8} & \textbf{9} & \textbf{10} & \textbf{11} & \textbf{} \\
\midrule
% cRT \cite{shi2021overcomingcatastrophicforgettingincremental}
% & 77.16 & 74.41 & 71.31 & 68.08 & 65.57 & 63.08 & 62.44 & 61.29 & 60.12 & 59.85 & 59.30 & 65.69 \\
% TOPIC ~\cite{cheraghian2021semanticawareknowledgedistillationfewshot}
% & 68.68 & 62.49 & 54.81 & 49.99 & 45.25 & 41.40 & 38.35 & 35.36 & 32.22 & 28.31 & 26.28 & 43.92 \\
% SPPR~\cite{zhu2021selfpromotedprototyperefinementfewshot}
% & 68.68 & 61.85 & 57.43 & 52.68 & 50.19 & 46.88 & 44.65 & 43.07 & 40.17 & 39.63 & 37.33 & 49.32 \\
% CEC ~\cite{zhang2021fewshotincrementallearningcontinually}
% & 75.85 & 71.94 & 68.50 & 63.50 & 62.43 & 58.27 & 57.73 & 55.81 & 54.83 & 53.52 & 52.28 & 61.33 \\
% F2M ~\cite{shi2021overcomingcatastrophicforgettingincremental}
% & 77.13 & 73.92 & 70.27 & 66.37 & 64.34 & 61.69 & 60.52 & 59.38 & 57.15 & 56.94 & 55.89 & 63.96 \\
% Entropy-Reg~\cite{liu2022fewshotclassincrementallearningentropyregularized}
% & 75.90 & 72.14 & 68.64 & 63.76 & 62.58 & 59.11 & 57.82 & 55.89 & 54.92 & 53.58 & 52.39 & 61.52 \\
% LIMIT ~\cite{10.1109/TPAMI.2022.3200865}
% & 75.89 & 73.55 & 71.99 & 68.14 & 67.42 & 63.61 & 62.40 & 61.35 & 59.91 & 58.66 & 57.41 & 65.48 \\
% % MetaFSCIL~\cite{Chi_2022_CVPR}
% % & 75.90 & 72.41 & 68.78 & 64.78 & 62.96 & 59.99 & 58.30 & 56.85 & 54.78 & 53.82 & 52.64 & 61.93 \\
% %\midrule
% % FSLL~\cite{mazumder2021fewshotlifelonglearning}
% % & 72.77 & 69.33 & 65.51 & 62.66 & 61.10 & 58.65 & 57.78 & 57.26 & 55.59 & 55.39 & 54.21 & 60.93 \\
% SoftNet~\cite{softnet}
% & 78.07 & 74.58 & 71.37 & 67.54 & 65.37 & 62.60 & 61.07 & 59.37 & 57.53 & 57.21 & 56.75 & 64.68 \\
% \midrule
ALICE~\cite{peng2022few} & 67.63 & 58.42 & 42.51 & 37.58 & 55.33 & 36.24 & 55.90 & 49.83 & 48.47 & \textbf{65.97} & 58.00 & 52.35 \\
FACT~\cite{zhou2022forward} & 65.02 & 62.23 & 58.92 & 55.14 & 54.41 & 51.97 & 51.50 & 49.02 & 49.25 & 48.53 & 47.85 & 53.99 \\
WaRP~\cite{kim2023warping} & 77.65 & \underline{74.80} & \underline{71.60} & \underline{67.44} & 64.92 & 61.88 & 60.94 & 58.75 & 55.86 & 53.66 & 49.91 & 63.40 \\
NC-FSCIL \cite{yang2023neural} & \underline{78.42} & 71.08 & 67.41 & 63.49 & 59.91 & 57.61 & 55.29 & 54.39 & 51.23 & 49.13 & 48.02 & 59.63 \\
% Comp \cite{comp-fscil} & 80.94 & 77.51 & 74.34 & 71.00 & 68.77 & 66.41 & 64.85 & 63.92 & 62.12 & 62.10 & 61.17 & 68.47 \\
% Approx \cite{approx}  \\
% C-FSCIL \cite{hersche2022constrainedfewshotclassincrementallearning} & 75.00 & 68.02 & 63.56 & 59.90 & 57.08 & 53.74 & 52.25 & 50.44 & 48.33 & 48.22 & 47.34 & 56.72 \\
OrCo \cite{ahmed2024orco} & 73.01 & 65.70 & 63.64 & 61.94 & 60.45 & 59.22 & 58.41 & 57.59 & 55.84 & 56.44 & 56.11 & 60.76 \\
Comp \cite{zou2024compositional} & 78.39 & 74.51 & 71.17 & 67.41 & \underline{65.02} & \underline{63.01} & \underline{62.38} & \underline{61.39} & \underline{59.74} & 59.74 & \underline{59.11} & \underline{65.62} \\
\midrule
\textbf{Ours}
& \textbf{82.25} & \textbf{78.82} & \textbf{76.76} & \textbf{73.91} & \textbf{72.11} & \textbf{69.98} & \textbf{69.12} & \textbf{67.96} & \textbf{66.34} & \underline{64.27} & \textbf{62.51} & \textbf{71.28} \\
\bottomrule
\end{tabular}%
}
\label{tab:new_results_11_sessions}
\end{table*}

\begin{table*}[t]
\centering
\caption{Architecture study of our method on CIFAR-100 FSCIL (5-way 5-shot). We report classification accuracy (\%) using different backbone networks, including CNNx4, CNNx8, CNNx16, ResNet-18, ResNet-50, and ResNeXt-29. Best (per column) is in \textbf{bold}, second best is \underline{underlined}.}
\small
\setlength{\tabcolsep}{2.8pt} 
\renewcommand{\arraystretch}{1.15}
\begin{tabular}{lccccccccccc}
\toprule
\textbf{Backbone} &
\multicolumn{9}{c}{\textbf{Sessions (Accuracy \%)}} &
\textbf{Avg (\%)} \\
\cmidrule(lr){2-10}\cmidrule(lr){11-11}
 & \textbf{1} & \textbf{2} & \textbf{3} & \textbf{4} & \textbf{5} & \textbf{6} & \textbf{7} & \textbf{8} & \textbf{9} & \textbf{} \\
\midrule

\multicolumn{11}{c}{\textit{Without pretraining}} \\
CNNx4
& 61.92 & 58.75 & 56.13 & 53.72 & 50.93 & 48.62 & 45.62 & 43.95 & 42.03 & 51.30 \\
CNNx8
& 65.30 & 61.91 & 59.14 & 56.52 & 54.10 & 51.78 & 49.21 & 47.58 & 45.89 & 54.60 \\
CNNx16
& 68.05 & 64.43 & 61.76 & 58.93 & 56.13 & 53.82 & 51.18 & 49.44 & 47.50 & 56.80 \\
%ResNet-12
%& 79.72 & 75.89 & 72.50 & 69.36 & 66.45 & 64.09 & 61.22 & 59.39 & 57.37 & 67.33 \\
ResNet-18
& 82.50 & 77.22 & 72.69 & 70.55 & 67.76 & 64.40 & \underline{62.76} & \underline{60.31} & \underline{58.52}
& 68.52 \\
ResNet-50
% & \textbf{86.05} & \textbf{81.35} & \textbf{77.14} & \textbf{73.41} & \textbf{69.64} & \textbf{66.24} & \textbf{62.93} & \textbf{60.33} & \underline{57.70} & \textbf{70.53} \\
& \textbf{86.83} & \textbf{82.06} & \underline{77.76} & \underline{73.51} & \underline{69.45} & \underline{65.94} & 62.47 & 60.26 & 57.84 & \underline{70.68} \\
ResNeXt-29
& \underline{85.30} & \underline{81.55} & \textbf{78.23} & \textbf{74.73} & \textbf{71.50} & \textbf{68.75} & \textbf{65.53} & \textbf{63.12} & \textbf{61.11} & \textbf{72.20} \\
% WRN-10-10
% & 0.00 & 0.00 & 0.00 & 0.00 & 0.00 & 0.00 & 0.00 & 0.00 & 0.00 & 0.00 \\

% \midrule
% \multicolumn{11}{c}{\textit{With pretraining}} \\
% ResNet-18 (ImageNet)
% & 87.30 & 83.09 & 79.81 & 76.59 & 73.83 & 71.74 & 69.99 & 67.72 & 65.44 & 75.06 \\
%  ViT (ImageNet)
%  & \textbf{89.80} & \textbf{87.57} & \textbf{87.07} & \textbf{85.27} & \textbf{82.16} & \textbf{80.58} & \textbf{78.84} & \textbf{78.03} & \textbf{77.08} & \textbf{82.93} \\

\bottomrule
\end{tabular}
\label{tab:architecture_ablation}
\end{table*}

\section{Cross-Domain Mini-ImageNet to Caltech-101}
\label{app:cross_domain_caltech}
To further evaluate the robustness of the cross-domain results presented in Section~\ref{sec:cross_domain_results}, we conduct a second experiment using a different cross-domain dataset pair. Specifically, the base session is trained on 60 classes from Mini-ImageNet, while each subsequent session introduces 10 novel classes from Caltech-101 using only five examples per class. This setting is particularly challenging because the model must not only learn new classes from limited supervision, but also adapt to a different domain from that observed during base training. Results are reported in Table~\ref{tab:cross_domain_caltech}.

The results show that SBP remains robust under this larger distribution shift. SBP achieves the highest average accuracy over the complete sequence, reaching 64.87\%, compared with 62.64\% for the next-best method, ALICE. SBP also achieves the highest final-session accuracy of 58.77\%, exceeding WaRP, the second-best method at the final session, by 4.38\%.

The early sessions further illustrate the difficulty of the domain transition. ALICE initially performs slightly better than our method, achieving 71.60\% and 68.90\% at Sessions~1 and~2, compared with 70.34\% and 67.88\% for SBP. However, SBP overtakes ALICE by Session~3 and remains the best-performing method for every subsequent session. By Session~10, the gap in performance has grown to 5.77\% over ALICE. This pattern suggests that the benefit of SBP becomes more pronounced as additional classes from the target domain are introduced and interference accumulates across incremental updates.

The results also show that strong base-session performance alone is not sufficient for robust cross-domain FSCIL. ADBS obtains the highest base-session accuracy at 82.10\%, but its performance decreases rapidly after the domain shift, reaching only 26.60\% by Session~10. In contrast, SBP starts from a lower base accuracy of 77.33\% but retains 58.77\% after all incremental sessions. This represents a much smaller degradation across the sequence and highlights the importance of preserving previously learned representations while still allowing sufficient capacity to adapt to the target domain.

WaRP exhibits comparatively stable performance after the domain shift, declining from 64.41\% at Session~1 to 54.39\% at Session~10. WaRP also remains comparatively stable after the domain shift, but SBP maintains higher accuracy throughout the incremental sequence, with a 4.38\% advantage at the final session.

Overall, these results reinforce the findings from Section~\ref{sec:cross_domain_results}, showing that SBP is not limited to settings where base and incremental classes originate from the same data distribution. SBP shows strong performance even when incremental classes are drawn from a different dataset and domain, where the model must balance retention of the source-domain representation with adaptation to novel target-domain classes from very limited data. %Together, the cross-domain experiments indicate that explicitly controlling which parameters remain stable and which remain available for adaptation provides a useful mechanism for maintaining performance under both class-incremental and distribution-shift settings.

\begin{table*}[!t]
\centering
\caption{Cross-Domain FSCIL (MiniImageNet $\rightarrow$ Caltech-101) classification accuracy (\%) with a ResNet-18 backbone. Best (per column) is in \textbf{bold}, second best is \underline{underlined}.}
\footnotesize
\setlength{\tabcolsep}{3.0pt}
\renewcommand{\arraystretch}{1.15}
\resizebox{\textwidth}{!}{%
\begin{tabular}{lcccccccccccc}
\toprule
\textbf{Method} &
\multicolumn{11}{c}{\textbf{Sessions (Accuracy \%)}} &
\textbf{Avg (\%)} \\
\cmidrule(lr){2-12}
& \textbf{0} & \textbf{1} & \textbf{2} & \textbf{3} & \textbf{4} &
\textbf{5} & \textbf{6} & \textbf{7} & \textbf{8} & \textbf{9} &
\textbf{10} & \\
\midrule

ALICE \cite{peng2022few}
& 76.50 & \textbf{71.60} & \textbf{68.90} & \underline{66.40}
& \underline{63.70} & \underline{61.70} & \underline{59.50}
& \underline{57.60} & 55.80 & 54.30 & 53.00
& \underline{62.64} \\

OrCo \cite{ahmed2024orco}
& 76.50 & 63.60 & 61.40 & 59.20 & 58.00 & 57.00
& 55.40 & 54.50 & 53.10 & 51.80 & 50.30
& 58.25 \\

ADBS \cite{li2025adaptive}
& \textbf{82.10} & 68.30 & 61.60 & 50.70 & 44.20 & 41.60
& 37.80 & 33.50 & 30.20 & 28.90 & 26.60
& 45.95 \\

WaRP \cite{kim2023warping}
& 69.41 & 64.41 & 63.15 & 61.86 & 60.06 & 58.70
& 57.68 & 57.22 & \underline{56.11} & \underline{55.19}
& \underline{54.39} & 59.83 \\

\midrule
\textbf{Ours}
& \underline{77.33} & \underline{70.34} & \underline{67.88}
& \textbf{66.70} & \textbf{65.98} & \textbf{64.27}
& \textbf{62.86} & \textbf{60.47} & \textbf{59.76}
& \textbf{59.21} & \textbf{58.77}
& \textbf{64.87} \\

\bottomrule
\end{tabular}
}
\label{tab:cross_domain_caltech}
\end{table*}

\section{Long-Horizon Implementation Details and ViT Results}
\label{app:vit_long_horizon}

\begin{table*}[!t]
\centering
\caption{
Long-horizon FSCIL on ImageNet-1K. Session~0 denotes the base session, followed
by 80 incremental sessions. ADBS, OrCo, WaRP, and SBP (ResNet-18) use the same
ResNet-18 backbone and are directly comparable. SBP (ViT-B/16) is reported
separately as an additional backbone experiment and is not included in the
best/second-best highlighting of the ResNet-18 comparison.
}
\footnotesize
\setlength{\tabcolsep}{3.5pt}
\renewcommand{\arraystretch}{1.15}
\resizebox{\textwidth}{!}{%
\begin{tabular}{lccccccccccc}
\toprule
\textbf{Method} &
\multicolumn{11}{c}{\textbf{Sessions (Accuracy \%)}} \\
\cmidrule(lr){2-12}
& \textbf{0} & \textbf{1} & \textbf{5} & \textbf{10}
& \textbf{20} & \textbf{30} & \textbf{40} & \textbf{50}
& \textbf{60} & \textbf{70} & \textbf{80} \\
\midrule

\multicolumn{12}{l}{\textit{ResNet-18 backbone}} \\

ADBS \cite{li2025adaptive}
& 61.50
& 57.10
& 47.20
& 39.20
& 27.40
& 22.20
& 18.80
& 17.20
& 16.00
& 14.90
& 14.50 \\

OrCo \cite{ahmed2024orco}
& 27.00
& 24.80
& 24.90
& 25.10
& 24.30
& 23.20
& 21.90
& 21.00
& 20.00
& 19.30
& \underline{18.40} \\

WaRP \cite{kim2023warping}
& \underline{63.20}
& \underline{61.90}
& \underline{60.40}
& \underline{58.60}
& \textbf{54.90}
& \textbf{51.40}
& \underline{47.80}
& \underline{42.10}
& \underline{33.30}
& \underline{20.00}
& 12.90 \\

\textbf{SBP (ResNet-18)}
& \textbf{72.45}
& \textbf{70.09}
& \textbf{64.99}
& \textbf{61.85}
& \underline{54.26}
& \underline{50.96}
& \textbf{47.93}
& \textbf{45.19}
& \textbf{43.01}
& \textbf{40.83}
& \textbf{39.08} \\

\midrule
\multicolumn{12}{l}{\textit{Additional backbone experiment}} \\

SBP (ViT-B/16)
& 73.56
& 70.19
& 65.03
& 63.28
& 58.85
& 55.78
& 53.08
& 50.77
& 48.93
& 46.85
& 45.07 \\

\bottomrule
\end{tabular}
}
\label{tab:vit_long_horizon_with_vit}
\end{table*}

% The long-horizon results reveal differences that are not apparent over shorter FSCIL sequences. WaRP remains competitive during the early part of the stream and slightly outperforms the ResNet-18 SBP model at Sessions~20 and~30. However, its performance degrades substantially thereafter, falling from 51.40\% at Session~30 to 12.90\% at Session~80. In contrast, SBP retains 39.08\% accuracy after all 80 incremental sessions. ADBS and OrCo similarly reach only 14.50\% and 18.40\% accuracy, respectively, at the final session. These results reinforce the observation from Section~\ref{sec:long_horizon} that performance over short FSCIL sequences does not necessarily predict behavior over substantially longer learning horizons, where accumulated interference becomes increasingly important.

% For the long-horizon experiment, we additionally evaluate SBP with a ViT-B/16 backbone to examine whether its long-horizon behavior is specific to ResNet-18. We use a ViT-B/16 with 224$\times$224 inputs and 768-dimensional representations. Importantly, the ViT is not externally pretrained: it is trained from scratch exclusively on the same 600 ImageNet-1K base classes used by the FSCIL protocol. We train it for 300 epochs on the base session. 

For the long-horizon ImageNet-1K experiment, we train the backbone using 600 classes for the base session and introduce the remaining 400 classes over 80 incremental sessions, each following a 5-way 5-shot protocol. The ResNet-18 backbone is trained from scratch using only the 600 base classes. We treat this base-training stage as the representation-learning stage for Session~0 and subsequently imprint the base-class prototypes without additional base-session optimization. During incremental learning, each session is trained for 75 iterations. We allocate 90\% of the network capacity to the base session and distribute the remaining 10\% uniformly across the 80 incremental sessions. Few-shot support examples are sampled independently for each session using a fixed random seed, and evaluation is performed on the ImageNet-1K validation set restricted to all classes observed up to the current session.

The ResNet-18 comparison shows that long-horizon rankings can differ substantially from those observed early in the sequence. WaRP slightly outperforms SBP at Sessions~20 and~30, but subsequently degrades from 51.40\% at Session~30 to 12.90\% at Session~80. In contrast, SBP (ResNet-18) retains 39.08\% accuracy after all 80 incremental sessions. 

We additionally evaluate SBP with a ViT-B/16 backbone to examine whether this long-horizon behavior is specific to ResNet-18. The ViT variant retains 45.07\% accuracy at Session~80, compared with 39.08\% for SBP (ResNet-18). Within SBP, the ViT variant retains 45.07\% accuracy at Session 80, compared with 39.08\% for the ResNet-18 variant. This experiment therefore provides an architecture-level robustness check rather than a direct comparison with the ResNet-18 baselines.

\section{Architecture Study} %\textcolor{blue}{Update text}}
\label{app:architectures}

We conduct an extensive architectural study on CIFAR-100 (same setup as in ablation study) to further validate the robustness and scalability of our method. Specifically, we test shallow convolutional networks (CNN-4, CNN-8, CNN-16) as well as %deeper 
residual networks (%ResNet-12, 
ResNet-18, ResNet-50) and ResNeXt-29 to demonstrate that SBP is not tied to a specific network topology and scales effectively with representational capacity. Table~\ref{tab:architecture_ablation} summarizes the per-increment and average accuracy for each backbone. We observe a strong positive correlation between network capacity and FSCIL performance. As we move from CNNx4 (51.30\% Avg.) to CNNx16 (56.80\% Avg.), our method consistently improves in per-increment accuracies. The transition to ResNet-based backbones results in a significant performance leap, with ResNet-50 achieving an average accuracy of 70.68\%, followed by ResNet-18 at 68.52\%. ResNeXt-29 further improves the average accuracy to 72.20\%, indicating that grouped-convolution architectures can provide additional capacity benefits under the same from-scratch FSCIL setting. 
% The pretrained results further show that SBP can also leverage stronger initial representations. ImageNet-pretrained ResNet-18 improves the average accuracy to 75.06\%. This result is reported separately because external pretraining changes the learning setting, but it indicates that SBP is compatible with stronger pretrained feature extractors. 
Across all tested architectures, SBP maintains a gradual accuracy decline across sessions rather than abrupt collapse, suggesting that its parameter-budgeting mechanism remains effective across backbone families and capacity scales.

% The pretrained results further show that SBP can also leverage stronger initial representations. ImageNet-pretrained ResNet-18 improves the average accuracy to 75.61\%, while the pretrained ViT achieves 82.93\%. These results are reported separately because external pretraining changes the learning setting, but they indicate that SBP is compatible with both convolutional and transformer-based feature extractors. Across all tested architectures, SBP maintains a gradual accuracy decline across sessions rather than abrupt collapse, suggesting that its parameter-budgeting mechanism remains effective across backbone families and capacity scales.
%We also report pretrained results separately, where ResNet-18 (ImageNet) reaches 75.61\% average accuracy and ViT (ImageNet) achieves 82.93\%, further showing that SBP can also benefit from stronger pretrained representations while preserving stable incremental performance.} Notably, across all tested backbones, our method maintains a smooth performance decline, avoiding catastrophic forgetting regardless of backbone size.

\begin{table*}[t]
\centering
\caption{Impact of base session capacity budget on CIFAR-100 FSCIL performance. %We report classification accuracy over various base session budgets for each increment. 
We vary the fraction of network capacity allocated to the base session and report per-session classification accuracy (\%) and average accuracy over all sessions. Best (per column) is in \textbf{bold}, second best is \underline{underlined}.}
\label{tab:budgets}
\small
\setlength{\tabcolsep}{2.8pt} 
\renewcommand{\arraystretch}{1.15}
\begin{tabular}{lccccccccccc}
\toprule
\textbf{Budget} &
\multicolumn{9}{c}{\textbf{\textbf{Sessions (Accuracy \%)}}} &
\textbf{Avg (\%)} \\
\cmidrule(lr){2-10}\cmidrule(lr){11-11}
 & \textbf{1} & \textbf{2} & \textbf{3} & \textbf{4} & \textbf{5} & \textbf{6} & \textbf{7} & \textbf{8} & \textbf{9} & \textbf{} \\
\midrule
50\% &
74.87 & 71.20 & 68.09 & 65.32 & 62.14 & 59.28 & 56.14 & 53.68 & 52.14 &
62.54 \\
75\% &
81.53 & 76.98 & 73.26 & 69.55 & 66.09 & 63.69 & 60.21 & 57.84 & 55.87 &
67.22 \\
80\% &
81.93 & 77.62 & 74.01 & 70.47 & 66.81 & 64.39 & 61.37 & 59.03 & 57.01 &
68.07 \\
85\% &
82.65 & 77.98 & 74.31 & 70.89 & 67.35 & 64.36 & 60.77 & 58.47 & \underline{57.84} &
68.29 \\
90\% &
82.32 & 77.95 & 74.17 & 70.75 & 67.06 & 64.73 & 61.73 & 59.37 & 56.89 &
68.33 \\
95\% &
\textbf{83.27} & \textbf{79.25} & \textbf{75.60} & \textbf{72.33} & \textbf{68.61} & \textbf{66.00} & \textbf{62.74} & \textbf{60.48} & \textbf{58.14} &
\textbf{69.60} \\
99\% &
\underline{83.25} & \underline{79.09} & \underline{74.97} & \underline{71.56} & \underline{68.08} & \underline{65.28} & \underline{62.01} & \underline{59.56} & 57.44 &
\underline{69.03} \\
\bottomrule
\end{tabular}
\end{table*}

\section{Effect of Base Session Budgets}
\label{app:budgets}

%We investigate the sensitivity of our method to the base session budget by scaling the base training duration while keeping all other incremental session hyperparameters fixed. 
We investigate the sensitivity of SBP to the base-session capacity budget by varying the fraction of network capacity allocated to the base session while keeping the training schedule and all incremental-session hyperparameters fixed. Specifically, we train the base session for 130 epochs and then run the same 8 incremental sessions (5-way 5-shot) as in the main experiments. The goal is to quantify how the base session budget influences the representation that supports efficient incremental adaptation.

Table~\ref{tab:budgets} reports the per-session classification accuracy for budgets of $\{50\%, 75\%, 80\%, 85\%, 90\%, 95\%, 99\%\}$. Overall, increasing the base session budget improves performance consistently across the incremental stream, with the largest gains appearing at the earliest increments. In particular, moving from $50\%$ to $75\%$ yields a substantial improvement in average accuracy (62.54\% $\rightarrow$ 67.22\%) and improves the final-session accuracy (52.14\% $\rightarrow$ 55.87\%, +3.73\%), indicating that %additional base compute significantly strengthens the initial feature extractor. 
allocating more capacity to the base session strengthens the initial representation. This is likely because assigning too many parameters to incremental sessions with only a few training samples leads to suboptimal results. 

Beyond 75\%, gains become more incremental: 80\% and 90\% improve the average accuracy to 68.07\% and 68.33\%, respectively, while 95\% achieves the best overall average accuracy at 69.60\%. Interestingly, 99\% is very close at 69.03\%, suggesting diminishing returns from further allocating additional capacity to the base session. We believe this is likely due to the model's backbone being largely frozen with a base budget of 99\%. Incremental sessions therefore have limited capacity to reshape global features, and performance becomes dominated by the quality of the base-session representation and the geometry it induces for prototype-based classification.

Our results suggest that moderate base session budgets (e.g., 75\%--90\%) already recover most of the attainable performance while preserving lower base training cost, whereas near-full budgets (95\%--99\%) offer only marginal additional gains in average accuracy. In settings where the base session is too expensive to train or must be trained quickly, reducing the base budget to 75\% can provide a strong accuracy--efficiency compromise, while still maintaining stable incremental behavior.

\section{Theoretical Motivation for Free Neuron Re-initialization in SBP}
\label{sec:theory_reset}

In Section~\ref{sec:sbp}, we introduce the epoch-wise re-initialization of free parameters ($\boldsymbol{\theta}_{\text{free}}$) as a core component of SBP. A critical challenge in parameter isolation is preventing unallocated capacity from becoming entangled with the current session's optimization. To formalize this, consider the network's representation at session $t$ as a linear superposition of three subspaces:
\begin{equation}
    f_t(x) = \mathbf{W}^{l, t}_{\mathrm{prev}} x + \mathbf{W}^{l, t}_{\mathrm{assigned}} x + \mathbf{W}^{l, t}_{\mathrm{free}} x,
\label{eq:superposition}
\end{equation}
where $\mathbf{W}_{\mathrm{prev}}$ and $\mathbf{W}_{\mathrm{assigned}}$ are the weights allocated to past and current tasks, respectively, and $\mathbf{W}_{\mathrm{free}}$ is the unallocated pool reserved for future sessions.

If the free parameters $\mathbf{W}_{\mathrm{free}}$ remain static during the optimization of task $t$, the greedy nature of backpropagation will cause $\mathbf{W}_{\mathrm{assigned}}$ to co-adapt to the specific noise realization present in the free block to minimize the task loss $\mathcal{L}_t$. This creates a latent coupling where the current task's performance becomes contingent on the state of future capacity. Consequently, when a subset of $\mathbf{W}_{\mathrm{free}}$ is promoted to $\mathbf{W}_{\mathrm{assigned}}$ in session $t+1$, it no longer follows a clean, unbiased distribution, but is instead biased by the training history of previous sessions.

\begin{figure}[t]
    \centering
    \begin{subfigure}[t]{0.99\textwidth}
        \centering
        \includegraphics[width=\linewidth]{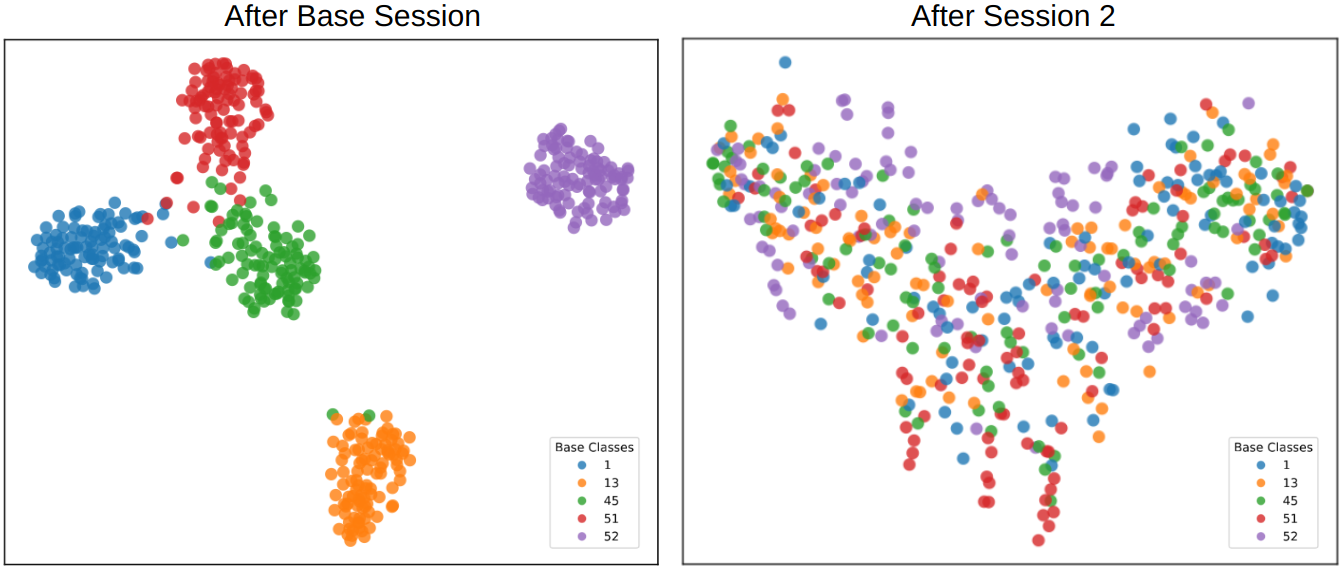}
        \caption{\textit{Without} free-parameter re-initialization.}
    \end{subfigure}
    %\hfill
    
    \begin{subfigure}[t]{0.99\textwidth}
        \centering
        \includegraphics[width=\linewidth]{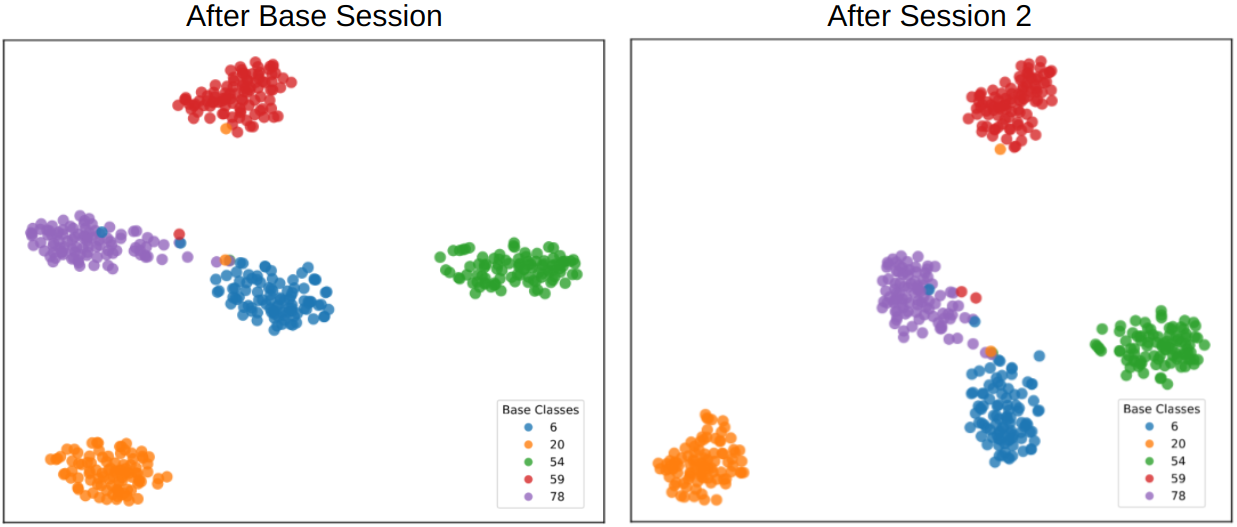}
        \caption{\textit{With} epoch-wise free-parameter re-initialization.}
    \end{subfigure}
    \caption{%Evolution of base class features with and without SBP's free-parameter re-initialization. Each point represents a sample, with color indicating its class. Without re-initialization, classes form distinct clusters after the base session, but updating the implicitly co-adapted ``free'' parameters in Session 2 causes a catastrophic collapse. With re-initialization, the features remain tightly clustered and well-separated after Session 2, showing invariance to changes in the unallocated capacity.
    Evolution of base-class features with and without SBP's free-parameter reinitialization. Each point represents a sample, with color indicating its class. Without reinitialization, base classes are initially separated after Session 1, but collapse after Session 2 once the implicitly co-adapted free parameters are reassigned and updated. With epoch-wise reinitialization, the feature clusters remain compact and well separated after Session 2, indicating that the learned representation is invariant to changes in the free subspace.}
    \label{fig:tsne_with_free_init}
\end{figure}

To enforce strict temporal decoupling, SBP resets the free partition $\boldsymbol{\theta}^{l}_{\mathrm{free}}$ at the start of every epoch to independent draws from a zero-mean distribution $\mathcal{D}$ (e.g., Kaiming Uniform). By treating the free parameters as a zero-mean random variable, we ensure that their expected contribution to the output vanishes over the course of optimization:
\begin{equation}
    \mathbb{E}_{\mathbf{W}_{\mathrm{free}} \sim \mathcal{D}} \left[ \mathbf{W}^{l, t}_{\mathrm{free}} x \right] = 0.
\end{equation}
This formulation forces the optimizer to find a solution for $\mathbf{W}_{\mathrm{assigned}}$ that is invariant to the values in the free subspace. This mechanism provides two key benefits: (i) the current task learns robust representations that do not consume unused capacity, and (ii) future tasks inherit an unbiased initialization that is independent of the previous training history and optimizer states.

Importantly, free-parameter reinitialization is not merely an implementation detail, but the mechanism that temporally decouples current learning from future capacity. Without reinitialization, unused parameters can remain fixed during training and become a stable, implicit part of the current task's solution, even though they receive no gradients. This creates hidden co-adaptation between \(\theta_{\mathrm{assigned}}\) and \(\theta_{\mathrm{free}}\), so that when free parameters are later reassigned to future sessions, changing them can destroy representations learned earlier. By reinitializing \(\theta_{\mathrm{free}}\) at every epoch, SBP turns the free subspace into a non-stationary nuisance component. The optimizer is therefore forced to learn current-task representations that are invariant to this subspace and depend only on the assigned parameters. As a result, current tasks do not consume future capacity, and future tasks inherit parameters that remain statistically independent of previous-session optimization. This explains why SBP can use all parameters at inference time without explicit forward masking: the learned representation has been trained to ignore the free subspace. The severe collapse observed when removing free-parameter reinitialization further confirms that this mechanism is essential for preventing latent co-adaptation and preserving reusable capacity across incremental sessions.

%\textcolor{blue}{Crucially, this expectation-based argument translates directly to deterministic test-time behavior without the need for explicit forward masking. Because the optimizer is forced to learn solutions that are invariant to the non-stationary free subspace during training, the learned network inherently ignores these parameters. This is directly analogous to dropout-style robustness, where training-time stochasticity yields deterministic yet invariant solutions at test time. Consequently, even though inference is deterministic and the free parameters contain random initialization values, their contribution to the final network output remains negligible, effectively preventing the injection of stochastic noise into the predictions.}

\section{%Analysis of the Feature Space Without Free-Parameter Re-initialization
Feature-Space Analysis of Free-Parameter Reinitialization}
\label{sec:tsne_analysis}

In our ablation studies (see Section \ref{sec:ablation_study}), we observed that removing the epoch-wise re-initialization of free parameters (\textit{w/o Free-Init}) results in an abrupt model failure, causing accuracy to plummet to $\sim1.56\%$ by the final session. Here, we provide a detailed diagnosis of this failure mode and visually demonstrate why SBP's temporal decoupling is essential. 

To illustrate this, we visualize the feature embeddings of five base classes from the CIFAR-100 dataset, using t-SNE. As shown in Figure~\ref{fig:tsne_with_free_init}(a), without free-parameter re-initialization, the base classes form distinct, well-separated clusters after Session 1. However, immediately after Session 2, the feature space collapses; the previously distinct clusters scatter and overlap entirely, visually confirming catastrophic forgetting. Conversely, Figure~\ref{fig:tsne_with_free_init}(b) shows the feature space evolution when utilizing the full SBP method with epoch-wise re-initialization. Because the free parameters remain non-stationary during the base session, the optimizer is strictly prevented from utilizing them to separate the base classes. Consequently, when these parameters are assigned and updated in Session 2, the base representations remain untouched. This explains why free-parameter reinitialization is critical for SBP: it prevents latent co-adaptation, preserves clean capacity for future sessions, and enables controlled plasticity without requiring explicit forward masking at inference time.

\end{document}